\documentclass[preprint,12pt]{elsarticle}
\usepackage[pdfborder={0 0 0}]{hyperref}
\usepackage{amsthm,amsmath,amsfonts,latexsym,amssymb}
\usepackage{graphics,fancyhdr,graphicx}
\usepackage[ruled,linesnumbered]{algorithm2e}
\usepackage[margin=1in]{geometry}
\usepackage{multirow,booktabs}
\usepackage[table]{xcolor}
\usepackage{listings}
\usepackage{tabularx}
\usepackage{placeins}
\usepackage{comment}
\usepackage{lineno}
\usepackage{lscape}
\usepackage{bm}
\usepackage{hyperref}
\usepackage{mathrsfs}
\usepackage{arydshln}
\usepackage{appendix}
\usepackage{comment}
\definecolor{custom-blue}{RGB}{3,69,173}
\hypersetup{
    colorlinks = true,
    urlcolor   = blue,
    citecolor  = custom-blue,
}
\biboptions{numbers,sort&compress}
\definecolor{listinggray}{gray}{0.9}
\definecolor{lbcolor}{rgb}{0.9,0.9,0.9}
\definecolor{Darkgreen}{RGB}{0,100,0}
\usepackage{caption}

\usepackage{float}
\usepackage{booktabs}
\usepackage{multirow}
\usepackage{tabularray}
\usepackage{mathtools}
\usepackage{subcaption}
\usepackage{enumitem}
\newcommand{\ie}{\textit{i.e.}}
\newcommand{\eg}{\textit{e.g.}}

\newtheoremstyle{remarkstyle}  
  {5pt}                        
  {5pt}                        
  {}                           
  {}                           
  {\bfseries}                  
  {.}                          
  { }                          
  {}                           
\newtheorem{theorem}{Theorem}[section]

\begin{document}

\makeatletter
\def\ps@pprintTitle{%
  \let\@oddhead\@empty
  \let\@evenhead\@empty
  \let\@oddfoot\@empty
  \let\@evenfoot\@oddfoot
}
\makeatother

\abovedisplayskip=6.0pt
\belowdisplayskip=6.0pt

\begin{frontmatter}

\title{Separable Physics-Informed DeepONet: Breaking the Curse of Dimensionality in Physics-Informed Machine Learning}

\author[1]{Luis Mandl}
\author[2]{Somdatta Goswami\corref{cor1}}
\ead{sgoswam4@jhu.edu}
\author[1]{Lena Lambers}
\author[1]{Tim Ricken}

\address[1]{Institute of Structural Mechanics and Dynamics in Aerospace Engineering, Faculty of Aerospace Engineering and Geodesy, University of Stuttgart, Germany.}
\address[2]{Department of Civil and Systems Engineering, Johns Hopkins University, U.S.A.}

\cortext[cor1]{Corresponding author.}

\begin{abstract}
\noindent
The deep operator network (DeepONet) has shown remarkable potential in solving partial differential equations (PDEs) by mapping between infinite-dimensional function spaces using labeled datasets. However, in scenarios lacking labeled data, the physics-informed DeepONet (PI-DeepONet) approach, which utilizes the residual loss of the governing PDE to optimize the network parameters, faces significant computational challenges, particularly due to the curse of dimensionality. This limitation has hindered its application to high-dimensional problems, making even standard 3D spatial with 1D temporal problems computationally prohibitive. Additionally, the computational requirement increases exponentially with the
discretization density of the domain. To address these challenges and enhance scalability for high-dimensional PDEs, we introduce the Separable physics-informed DeepONet (Sep-PI-DeepONet). This framework employs a factorization technique, utilizing sub-networks for individual one-dimensional coordinates, thereby reducing the number of forward passes and the size of the Jacobian matrix required for gradient computations. By incorporating forward-mode automatic differentiation (AD), we further optimize computational efficiency, achieving linear scaling of computational cost with discretization density and dimensionality, making our approach highly suitable for high-dimensional PDEs. We demonstrate the effectiveness of Sep-PI-DeepONet through three benchmark PDE models: the viscous Burgers' equation, Biot's consolidation theory, and a parametrized heat equation. Our framework maintains accuracy comparable to the conventional PI-DeepONet while reducing training time by two orders of magnitude. Notably, for the heat equation solved as a 4D problem, the conventional PI-DeepONet was computationally infeasible (estimated 289.35 hours), while the Sep-PI-DeepONet completed training in just 2.5 hours. These results underscore the potential of Sep-PI-DeepONet in efficiently solving complex, high-dimensional PDEs, marking a significant advancement in physics-informed machine learning.
\end{abstract}
\end{frontmatter}

\section{Introduction}

Neural operator learning, which employs deep neural networks (DNNs) to learn mappings between infinite-dimensional function spaces, has recently gained significant attention, particularly for its applications in learning partial differential equations (PDEs). A classical solution operator learning task involves learning mappings across various scenarios, such as different domain geometries, PDE parameters, and initial and boundary conditions (I/BCs), to the solution of the underlying PDE system. Currently, there are numerous neural operators, which can be categorized into meta-architectures, such as deep operator networks (DeepONet)~\cite{lu2021learning}, and operators based on integral transforms, including the Fourier neural operator (FNO)~\cite{li2021fourier}, wavelet neural operator (WNO)~\cite{tripura2023wavelet}, graph kernel network (GKN)~\cite{anandkumar2020neural}, and Laplace neural operator (LNO)~\cite{cao2023lno}, among others.

Among the various neural operators developed, in this work, we consider the deep operator network (DeepONet) for its architectural flexibility. Its architecture, inspired by the universal approximation theorem for operators~\cite{chen1995universal}, consists of a branch network encoding the varying input functions and a trunk network encoding spatiotemporal coordinates. While traditional DeepONet implementations rely on extensive labeled datasets, which can be computationally expensive or experimentally challenging to obtain, the physics-informed DeepONet (PI-DeepONet) offers an alternative by embedding physical laws into the learning process. Deep neural operators are promising methods to obtain digital twins with generalized and near real-time inference capabilities~\cite{Kobayashi2024}. Research has focused on adapting the fundamental deep neural operator architecture to address specific challenges across various domains. For instance, Geom-DeepONet incorporates parameterized 3D geometries~\cite{He2024}, while Sequential DeepONets employ recurrent architectures to capture time-dependent loading sequences~\cite{He2024b}. To enhance physical consistency, the Energy-Dissipative Evolutionary Deep Operator Neural Network enforces energy dissipation laws, enabling it to learn entire classes of partial differential equations rather than being limited to single equations with varying parameters and boundary conditions~\cite{Zhang2024}. Application-specific adaptations have also been developed, such as sequential learning and residual U-nets for additive manufacturing processes~\cite{Kushwaha2024}. Parallel to these applied developments, researchers are addressing fundamental theoretical challenges. Notable examples include the Phase-Field DeepONet, which learns pattern formation in gradient flows of free-energy functionals through physics-informed approaches~\cite{Li2023}. Additionally, comparative studies using the Poisson equation have evaluated the performance of classical versus physics-informed DeepONet in modeling spatial heat sources~\cite{Koric2023}.

The concept of physics-informed DeepONet (PI-DeepONet), introduced concurrently by Wang et al.~\cite{Wang2021} and Goswami et al.~\cite{goswami2022physics}, has remained largely unexplored in the scientific community since its inception, primarily due to the prohibitively high computational cost associated with the computation of gradient terms in the PDE. This challenge is more pronounced for large domains and high-dimensional systems, especially in cases involving parameterized problems that must account for variations in domain geometry, boundary conditions, and governing equation parameters. To understand the challenges of the conventional PI-DeepONet framework, consider an example of an $n^d$ discretization of a $d$-dimensional domain, with $n$ one-dimensional coordinates on each axis and $N$ input functions for the branch network. For each sample evaluated through the branch network, the trunk network evaluates $n^d$ forward passes of a single multi-layer perceptron (MLP). Furthermore, while computing the gradients using reverse-mode automatic differentiation (AD), the Jacobian matrix has a size of $kn^d \times kn^d$. The computational cost of evaluating this matrix increases exponentially with the increase in discretization density $n$ even for a standard 3D problem ($d=3$), restricting the scalability of the approach for solving high-dimensional PDEs. 

Recent efforts to enhance the scalability of physics-informed neural networks (PINNs) have yielded promising strategies that could potentially be adapted to PI-DeepONet. These include the investigation of the zero-coordinate shift to optimize calculations in reverse-mode AD~\cite{Leng2024}, as well as the utilization of factorizable coordinates and low-rank tensor decomposition to enable forward-mode AD~\cite{cho2023separable}. Low-rank tensor decomposition approaches, such as the Proper Generalized Decomposition (PGD) framework introduced by Chinesta et al.\cite{chinesta2013proper}, have gained attention for mitigating the curse of dimensionality in high-dimensional problems. While successfully applied in various domains\cite{vella2022pgd}, challenges in finding optimal low-rank approximations~\cite{de2008tensor} have led to advanced tensor approximation methods~\cite{billaud2014tensor} aimed at improving stability and efficiency.

\begin{figure}[hbt!]
    \includegraphics[width=\textwidth]{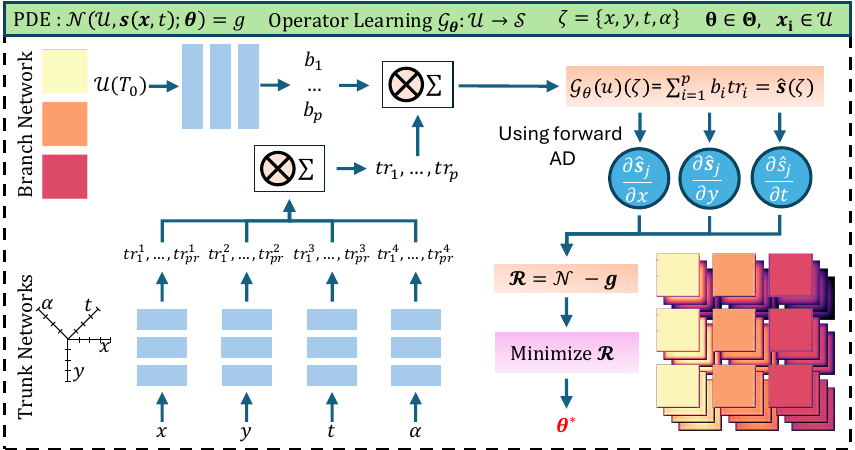}
    \caption{The framework of physics-informed separable DeepONet demonstrated for the parametrized heat equation example. Its central component is the outer product over the individual batches in the inputs followed by the summations indicated by $\bigotimes\sum$. This is done over the tensor rank $r$ in the trunk as well as the output over the hidden latent dimension $p$. Input to the trunk networks are factorizable coordinates and parameters. A detailed representation of the network structure including the batches is shown in Figure \ref{fig:sep_framework} in the Supplementary Materials. The framework employs forward-mode automatic differentiation to compute the derivative terms in the PDE.}
    \label{fig:separablePINO}
\end{figure}

These advancements in PINN optimization techniques suggest potential avenues for improving the computational efficiency of PI-DeepONet, potentially opening the door to its wider adoption and application in solving complex, high-dimensional PDEs. Drawing motivation from the recent developments in improving the optimization of PINNs, we aim to address the challenges of PI-DeepONet. To that end, we propose separable physics-informed DeepONet (Sep-PI-DeepONet) in this work which leverages the ideas of low-rank tensor decomposition to improve the computational efficiency of the vanilla PI-DeepONet framework drastically. The driving idea in our work is to gain efficiency by reducing the number of forward passes through trunk and branch networks and the size of the Jacobian matrix by leveraging the idea of separation of variables to solve PDEs. To that end, we propose the following modifications: \vspace{-10pt}
\begin{enumerate}
    \item \textbf{Factorized coordinates and separate sub-networks:} Instead of using a single MLP as the trunk network for all multidimensional coordinates, we employ factorized coordinates and separate sub-networks for each one-dimensional coordinate. Each sub-network processes its respective coordinate axis, and the final output is produced through an outer product and element-wise summation in the sense of a tensor rank approximation with rank $r$. This architecture reduces the number of trunk network propagation from $\mathcal{O}(Nn^d)$ to $\mathcal{O}(nd)$, for $N$ input functions. \vspace{-10pt}
    \item \textbf{Forward-mode AD for gradient terms:} To compute the gradients in the PDE, we use forward-mode AD, significantly reducing the computational cost of the Jacobian matrix. In this separated approach, the Jacobian matrix is of size $nd \times kn^d$ and requires $\mathcal{O}(nd)$ evaluations with forward-mode AD, compared to the conventional $\mathcal{O}(kn^d)$ evaluations with reverse-mode AD. \vspace{-10pt}
    \item \textbf{Combining branch and trunk outputs:} The total number of passes through the branch network (which encodes the input function) and the trunk network (which defines the evaluation parameters) is decreased by combining all branch outputs and trunk outputs in each batch. This is done by using an outer product followed by a summation over the hidden dimension (denoted as $p$) (\textit{einsum} operation), effectively acting as factorized inputs over the trunk and branch networks. This method further reduces the computational cost for forward passes of the trunk network from being proportional to $\mathcal{O}(kn^d)$ to being proportional to $\mathcal{O}(nd)$. \vspace{-10pt}
\end{enumerate}
These modifications will allow the physics-informed version for DeepONet architecture to scale linearly with $n$ and hence, is amenable to high-dimensional PDEs. The proposed framework for Sep-PI-DeepONet is presented in Figure~\ref{fig:separablePINO}. The network structure with batch tracing is shown in Figure~\ref{fig:sep_framework} in the Supplementary Materials alongside Algorithm \ref{alg:schematic} to visualize the operations. In a parallel and independent study, Yu et al.~\cite{yu2024separable} developed a separable architecture for DeepONet, which shares similarities with but differs in approach to decompose the trunk network from the approach presented in this work. Both, our current research and the concurrent development~\cite{yu2024separable} contribute to the ongoing efforts to enhance physics-informed neural operator methods, albeit through distinct methodological innovations.

\section{Physics-Informed DeepONet}
\label{sec:methods}

The goal of data-driven operator learning is to learn a mapping between two infinite-dimensional spaces on a bounded open set $\Omega \subset \mathbb{R}^D$, given a finite number of input-output pairs. Let $\mathcal{U}$ and $\mathcal{S}$ be Banach spaces of vector-valued functions defined as:
\begin{align}
    &\mathcal{U} = \{\Omega; u: \mathcal{X} \to \mathbb{R}^{d_u}\}, \quad \mathcal{X}\subseteq \mathbb{R}^{d_x}\\
    &\mathcal{S} = \{\Omega; s: \mathcal{Y} \to \mathbb{R}^{d_s}\}, \quad \mathcal{Y}\subseteq \mathbb{R}^{d_y}, 
\end{align}
where $\mathcal{U}$ and $\mathcal{S}$ denote the set of input functions and the corresponding output functions, respectively. The operator learning task is defined as $\mathcal{G}: \mathcal{U} \to \mathcal{S}$. The objective is to approximate the nonlinear operator, $\mathcal{G}$,  via the following parametric mapping:
\begin{equation}
    \mathcal{G}: \mathcal{U} \times \mathbf{\Theta} \rightarrow \mathcal{S} \quad \text{or} \quad \mathcal{G}_{\boldsymbol{\theta}}: \mathcal{U} \rightarrow \mathcal{S}, \quad \boldsymbol{\theta} \in \mathbf{\Theta},
\end{equation}
where $\mathbf{\Theta}$ is a finite-dimensional parameter space. In the standard setting, the optimal parameters $\boldsymbol{\theta}^*$ are learned by training DeepONet with a set of labeled observations $\mathcal{D} = \left\{(u^{i}, s^{i})\right\}_{i=1}^N$, which contains $N$ pairs of input and output functions. When a physical system is described by PDEs, it involves multiple functions, such as the PDE solution, the forcing term, the initial condition, and the boundary conditions. We are typically interested in predicting one of these functions, which is the output of the solution operator (defined on the space $\mathcal{S}$), based on the varied forms of the other functions, \ie, the input functions in the space $\mathcal{U}$.

DeepONet is inspired by the universal approximation theorem of operators \cite{chen1995universal}. The architecture of vanilla DeepONet consists of two DNNs: the branch network, which encodes the input functions $\mathcal{U}$ at $m$ fixed sensor points $\{x_1, x_2, \dots, x_m\}$, and the trunk network, which encodes the information related to the spatiotemporal coordinates $\zeta = \{x_i, y_i, z_i, t_i\}$, where $i={1,2, \ldots, d_y}$ and $d_y = n_x \times n_y \times n_z \times n_t$ obtained using mesh grid operation carried out on the discretization of each dimensional axis. The solution operator for an input realization $u_1$ can be expressed as:
\begin{equation}\label{eq:output_DeepONet}
    \begin{split}
      \mathcal G_{\boldsymbol \theta}(u_1)(\zeta) &= \sum_{i = 1}^p b_i \cdot tr_i = \sum_{i = 1}^{p}b_i(u_{1}(x_1), u_{1}(x_2), \ldots, u_{1}(x_m))\cdot tr_i(\zeta),  
\end{split}
\end{equation}
where $\{b_1, b_2, \ldots, b_p\}$ are the output embeddings of the branch network and $\{tr_1, tr_2, \ldots, tr_p\}$ are the output embeddings of the trunk network. For simplicity, let us consider $n_x=n_y=n_z=n_t = n$, therefore $d_y = n^4$. In Equation~\eqref{eq:output_DeepONet}, $\boldsymbol{\theta} = \left(\mathbf{W}, \mathbf{b} \right)$ includes the trainable parameters (weights, $\mathbf{W}$, and biases, $\mathbf{b}$) of the networks. The loss function is computed by simulating the solution operator on $n^4$ locations for each of the $N$ samples. The optimized parameters of the network, $\boldsymbol{\theta}^*$, are obtained by minimizing a standard loss function ($\mathcal{L}_1$, \ie, $\vert\vert\bm{x}-\hat{\bm{x}}\vert\vert_1$ or $\mathcal{L}_2$, \ie, $\vert\vert\bm{x}-\hat{\bm{x}}\vert\vert_2$ with ground truth $\bm{x}$ and network prediction $\hat{\bm{x}}$) using a standard optimizer. Note that the extension of the separable framework can be transferred to the branch network as well but lies beyond the scope of this work.

The training architecture of DeepONet, as proposed in the original work by Lu et al.~\cite{lu2021learning}, was computationally intensive. This method required repeating the branch network entries $n^4$ times for each sample to perform the dot product operation, thereby obtaining the solution operator as shown in Equation~\eqref{eq:output_DeepONet}. Later, Lu et al.~\cite{Lu2022} introduced a more efficient training approach that evaluated the trunk network only once for all $n^4$ coordinates, utilizing the \textit{einsum} operation for computing the dot product. While this approach significantly reduced computational costs, it does not apply to PI-DeepONet.

In PI-DeepONet, computing the loss function requires calculating the solution's gradients, typically using reverse-mode AD. A critical requirement of this method is that the output size (the solution) and the input size (the coordinates in the trunk network) must match. This condition invalidates the efficient training strategy proposed in~\cite{Lu2022}, as the trunk network must be evaluated $Nn^4$ times for a solution of the same dimensionality to enable the reverse-mode AD algorithm for gradient computation. Consequently, the Jacobian matrix size is $Nn^4 \times Nn^4$. In this work, we aim to significantly reduce the computational cost of training the vanilla PI-DeepONet by introducing a separable architecture. For clarity on the architecture and the benefits of the separable framework, we consider a 4D problem, with 3D in space and 1D in time.

\subsection{Separable DeepONet}
\label{subsec:separableDeepONet}

Separable DeepONet (Sep-DeepONet) is inspired by the concept of separation of variables to solve PDEs. A similar implementation for the PINNs framework was shown in a recent work~\cite{cho2023separable}. Sep-DeepONet leverages the concept of factorizable coordinates and thus transfers this idea to function-function mappings. More specifically, a DeepONet represents a mapping from Banach space to Banach space, based on the universal approximation theorems for operators. For a 4D problem (3D in space and 1D in time), the Sep-DeepONet consists of 4 trunk networks, each of which takes an individual 1-dimensional coordinate component as input. Each trunk network $tr^j: \mathbb{R} \to \mathbb{R}^{pr}$ transforms each of the 1D co-ordinates of the $j$-th coordinate axis into a feature representation, where $r$ denotes the low-rank tensor decomposition and $p$ denotes the conventional latent representation of DeepONet.

For each dimension, we first sample $n$ one-dimensional coordinates, which are considered as input to the independent trunk networks that output an embedding matrix $ tr^j_{n,1}, tr^j_{n,2}, \ldots, tr^j_{n,pr}$ of size $n\times pr$, where $j = \{1, 2, 3, 4\}$. These embeddings are then reshaped to size $n \times p \times r$ before performing the outer product operation. We define the trunk output as a product of the 4 trunk networks written as:\[tr_{n_x, n_y, n_z, n_t,p}(\zeta) = \sum_{i=1}^{r} \left(\prod_{j=1}^{4} tr^j_{n,p,r}(\zeta)\right),\] and employ the \textit{einsum} operation to carry this out, resulting in an output of size $[n_x, n_y, n_z, n_t, p]$. The outer product operation merges the features, enabling the trunk network to produce outputs on a lattice grid with only $4n$ forward passes instead of $n^4$ in the vanilla DeepONet, considering $n_x = n_y = n_z = n_t =n$. Hence, the curse of dimensionality in sampling locations can be avoided using the separable framework. The branch network outputs embeddings of size $N \times p$. The solution operator is constructed using the \textit{einsum} operation, where the summation is carried out over the last dimension of the matrix on $p$, therefore resulting in a matrix of size $N \times n_x \times n_y \times n_z \times n_t$. 

Although this framework reduces the number of forward passes on the trunk network, it cannot directly leverage this computational advantage in the physics-informed version because reverse-mode AD cannot be employed to compute the gradients, which forms a significant part of the computational cost in PI-DeepONet. To that end, we explore the forward mode AD as discussed in the next section.

\subsection{Physics-Informed Separable DeepONet}
\label{subsec:pi_sep_deepOnet}

The physics-informed variant is achieved by combining data loss stemming from the initial conditions and the boundary conditions and a residual loss stemming from governing physical equations. The loss function in a physics-informed DeepONet is written as: 
 \begin{equation}\label{eq:lossterms}
     \mathcal{L}(\bm{\theta}) = \lambda_{pde}\mathcal{L}_{\rm physics}(\bm{\theta}) + \lambda_{ic}\mathcal{L}_{\rm initial}(\bm{\theta}) + \lambda_{bc}\mathcal{L}_{\rm boundary}(\bm{\theta}), 
 \end{equation}
where $\lambda_{pde},\lambda_{ic}$ and $\lambda_{bc}$ are weighting factors to penalize the corresponding loss contributions. While the weighting factors could be manually modulated through trials, an efficient way is to adaptively obtain them during the training process~\cite{kontolati2023influence}. Let us consider a differential operator, $\mathcal{N}(u, s) = g $ with parameters $u \in\mathcal{U}$, \eg, input functions, and the unknown solutions $s\in\mathcal{S}$ with $\mathcal{U}$ and $\mathcal{S}$ representing Banach spaces. Under approximation of a DeepONet with $\mathcal G_{\bm \theta}(u;\zeta)=s(u(\zeta))$. The three loss terms of Equation~\eqref{eq:lossterms} are defined as:
\begin{equation}\label{eq:outerproduct}
\begin{split}
    \mathcal{L}_{\rm physics} &= \frac{1}{N \times n^4}\sum_{i=1}^{N}\sum_{j=1}^{n^4}
    \left|
    \mathcal{N}(\bm{u}^{i}, \mathcal G_{\bm\theta}(\bm{u}^{i},\zeta_j)) -g \right|^2, \\
    \mathcal{L}_{\rm initial} &= \frac{1}{N\times N_{\text{initial}}}\sum_{i=1}^{N}\sum_{j=1}^{N_{\text{initial}}}
    \left|
   \mathcal G_{\bm{\theta}}(\bm{u}^{i})(\zeta_{j}) - \mathcal G(\bm{u}^{i})(\zeta_{j})
    \right|^2, \\
     \mathcal{L}_{\rm boundary} &= \frac{1}{N \times N_{\text{boundary}}}\sum_{i=1}^{N}\sum_{j=1}^{N_{\text{boundary}}}
    \left|
   \mathcal G_{\bm{\theta}}(\bm{u}^{i})(\zeta_{j}) - \mathcal G(\bm{u}^{i})(\zeta_{j})
    \right|^2, 
\end{split}
\end{equation}
where $N_{\text{initial}}$ and $N_{\text{boundary}}$ denote the total number of domain points on which the initial and the boundary conditions are defined, respectively.  Now, while writing out the $\mathcal{L}_{\rm physics}$ loss, one needs to compute the gradients of the solution with respect to the inputs of the trunk network using AD techniques~\cite{baydin2018}. The outer product shown in Equation~\eqref{eq:outerproduct} allows the merging of all the coordinate axes to define the solution on a lattice grid. However, while sampling trunk locations with $4$ coordinate axes, with $n$ points per axis scales with $\mathcal{O}(n^4)$ in vanilla PI-DeepONet ($n^4$ total points obtained using mesh grid of 4 axes), the separable approach scales with $\mathcal{O}(4n)$. Now, the sizes of the inputs to the trunk and the final solution are different, we cannot efficiently employ reverse-mode AD to compute the gradients. Instead, we employ forward-mode AD to compute the Jacobian of size $[4n, 4^n]$.
For $N$ input functions, the branch network considers $N$ forward passes. The computational effort for evaluating a vanilla PI-DeepONet is $\mathcal{O}(Nn^4)$, which is reduced to $\mathcal{O}(4t+N)$ in the separable architecture. Note that this calculation only considers network computation and not the subsequent outer products and summations, which are non-significant. Furthermore, computational memory can be saved as each trunk and branch input needs to be stored only once, instead of several instances to compute all computations (due to forward-mode AD).

In the Sep-PI-DeepONet for the Burgers' example, 1,000 input functions are passed as a single batch. This requires only 1,000 branch computations and a total of 100 trunk computations (50 each for $t$ and $x$) for the residual loss, 102 trunk computations (100 for $t$ and 2 for $x$) for the boundary conditions, and 102 trunk computations (1 for $t$ and 101 for $x$) for the initial conditions. The total computation cost is therefore $\mathcal{O}(1300)$. In contrast, a vanilla PI-DeepONet requires a total of 2,500,000 residual computations, 100,000 boundary computations, and 101,000 initial computations for both the trunk and branch networks, resulting in a total computation cost of $\mathcal{O}(2701000)$. This is 2,078 times more computationally expensive compared to the separable architecture. It is important to note that computing gradients using forward-mode AD has a lower memory footprint and is significantly faster than reverse-mode AD for problems with higher output dimensionality than input dimensionality.

The framework can be further improved by sampling random locations in the trunk network for every iteration, compared to using equidistant grid points. This will be addressed in our future work.

\section{Numerical Examples}
\label{sec:results}

To demonstrate the advantages and efficiency of the Sep-PI-DeepONet, we learn the solution operator for three diverse PDE models of increasing complexity and dimensionality. First, we consider the viscous Burgers' equation to highlight the framework's ability to handle nonlinearity. Our goal is to learn the solution operator that maps initial conditions to the spatio-temporal solution of the 1D Burgers’ equation. Second, we examine a PDE describing the consolidation of a fluid-saturated body using Biot's theory. Here, we aim to learn the solution operator that maps any loading function at the drained surface to the full spatio-temporal solution of a 1D column with a permeable top and impermeable bottom. This represents a coupled problem with two field variables. Finally, we explore the parameterized heat equation for a 2D plate. In this case, we learn the solution operator for the temporal evolution of the temperature field given an initial temperature field and thermal diffusivity. For the first two examples, $d=2$, while for the third example $d=4$. In all three examples, we obtain the results without using any labeled training data.

\begin{table}[h]
    \footnotesize
    \centering
    \begin{tabular}{lllcccc}
        \toprule
        \multirow{2}{*}{\textbf{Problem}} & \multirow{2}{*}{\textbf{Model}} & \multirow{2}{*}{\textbf{d}} & 
        \multicolumn{3}{c}{\textbf{Relative $\mathcal{L}_2$ error}} & \textbf{Run-time} \\
        & & & Mean & Min & Max & (ms/iter.)\\
        \midrule
        \multirow{2}{*}{Burgers'} Equation & Vanilla & \multirow{2}{*}{$2$} & $5.2\rm{e}\text{-}2$  & $1.4\rm{e}\text{-}2$& $2.9\rm{e}\text{-}1$&  136.6 \\
        & Separable (Ours) & & $6.2\rm{e}\text{-}2$ & $1.7\rm{e}\text{-}2$& $2.9\rm{e}\text{-}1$& \textbf{3.64} \\
        \midrule
        \multirow{2}{*}{Consolidation Biot's Theory } & Vanilla & \multirow{2}{*}{$2$} & $7.7\rm{e}\text{-}2$ & $1.2\rm{e}\text{-}2$& $3.4\rm{e}\text{-}1$& 169.43 \\
        & Separable (Ours) & & $7.8\rm{e}\text{-}2$ & $1.7\rm{e}\text{-}2$& $3.9\rm{e}\text{-}1$& \textbf{3.68} \\
        \midrule
        \multirow{2}{*}{Parameterized Heat Equation} & Vanilla & \multirow{2}{*}{$4$} & - & -& -& 10,416.7 \\
        & Separable (Ours) &  & $8.6\rm{e}\text{-}2$& $3.2\rm{e}\text{-}2$& $2.0\rm{e}\text{-}1$& \textbf{91.73} \\
        \midrule
        Poisson's Equation & Separable (Ours)& $3$& $4.7\rm{e}\text{-}2$& $2.2\rm{e}\text{-}2$& $1.8\rm{e}\text{-}1$ & 35.54 \\
        \bottomrule
    \end{tabular}
    \caption{Comparison of relative $\mathcal{L}_2$ error and run-time for all applications presented in this work. In this table, \textit{Vanilla} refers to the conventional PI-DeepONet \cite{Wang2021}, while \textit{Separable} denotes our proposed Sep-PI-DeepONet approach, and $d$ denotes the dimension of the problem. For both frameworks, the same discretization density is considered and they exhibit comparable accuracy; however, the separable framework demonstrates a significant computational advantage, \underline{requiring approximately two orders of magnitude less computing time.} Notably, we could not complete the full training of the vanilla PI-DeepONet for the parameterized heat equation due to its prohibitively high computational demands. In a last example, we included a convolutional neural network in the branch for a random source field for Poisson's equation showcasing the flexibility of the proposed method without comparing it to a vanilla variant.}
    \label{tab:results}
\end{table}

\begin{figure}[hbt!]
     \centering
     \begin{subfigure}[b]{0.495\textwidth}
         \centering
         \includegraphics[width=\textwidth]{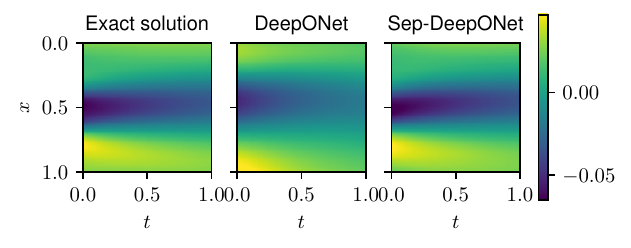}
         \caption{Burgers'} equation: Prediction of solution field after training for both networks for 83.69 s.
         \label{fig:time_compare_burgers}
     \end{subfigure}
     \hfill
     \begin{subfigure}[b]{0.495\textwidth}
         \centering
         \includegraphics[width=\textwidth]{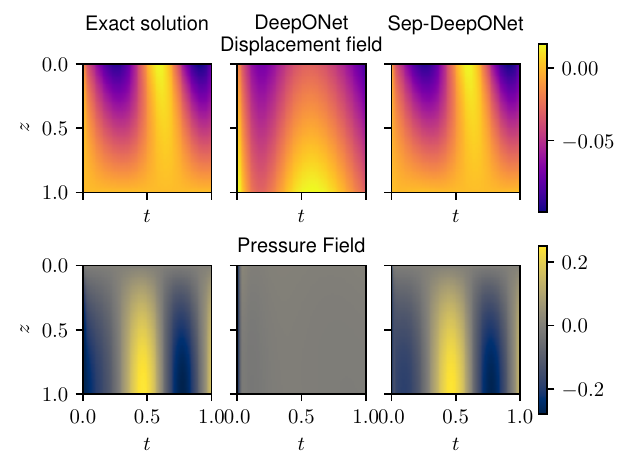}
         \caption{Consolidation Biot's Equation: Prediction after training both networks for 380.39 s.}
         \label{fig:time_compare_biot}
     \end{subfigure}

     \begin{subfigure}[b]{0.675\textwidth}
        \vspace{10pt}
         \centering
         \includegraphics[width=\textwidth]{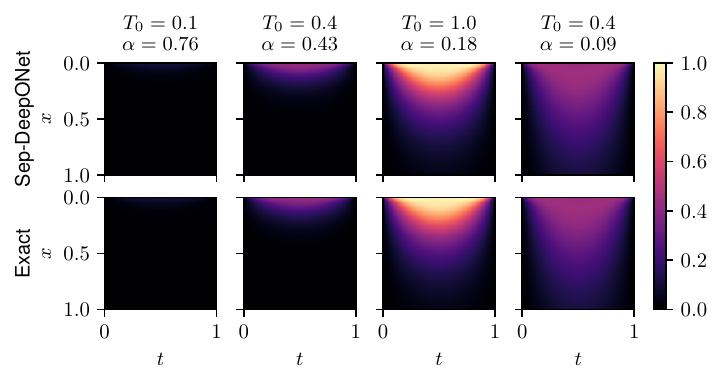}
         \caption{Parameterized heat equation along $y=0.5$ after convergence.}
         \label{fig:heat_overview}
     \end{subfigure}
     \hfill
    \begin{subfigure}[b]{0.295\textwidth}
        \vspace{10pt}
         \centering
         \includegraphics[width=\textwidth]{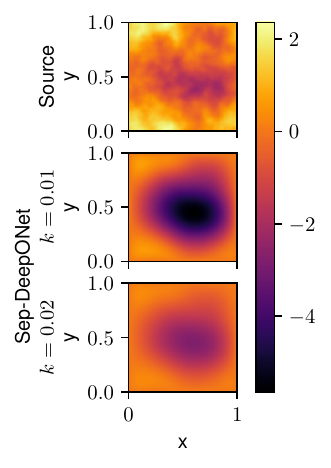}
         \caption{Poisson's equation after convergence at two diffusivities.}
         \label{fig:poisson_overview}
    \end{subfigure}
     
    \caption{Comparative results of vanilla PI-DeepONet and Sep-PI-DeepONet for all applications considered in this work, evaluated after fixed training time. (a) For the Burgers' equation (1D in space and 1D in time), after training for 83.69s: vanilla completed 600 iterations with $\mathcal{L}_2$ error of $3.82 \times 10^{-1}$, while separable completed 21,500 iterations with error $8.98 \times 10^{-2}$. (b) For the consolidation problem (1D in space and 1D in time), after training for 380.39s: vanilla completed 2,200 iterations achieving an error $4.11 \times 10^{-1}$ in the displacement field and an error of $9.36 \times 10^{-1}$ in the pressure field, while separable completed 95,500 iterations while achieving an error of $2.63 \times 10^{-2}$ in the displacement field and an error of $1.35 \times 10^{-1}$ in the pressure field. Networks for both problems have comparable trainable parameters. (c) For the parameterized heat equation (2D in space, 1D in parameter values, and 1D in time), the separable architecture was evaluated after convergence (approx. 2.5h training time). The training could not be completed with the vanilla framework as each iteration required 10,416.7 mili-seconds (approx. 289.35h training time) (see Table~\ref{tab:results}). (d) The separable framework is easily transferable to other inputs such as random sources for Poisson's equation and can be combined with other architectures, \eg, a convolutional neural network as a branch network.}
    \label{fig:time_compare}
\end{figure}

Figure \ref{fig:time_compare} illustrates the performance of the proposed Sep-PI-DeepONet in comparison to vanilla PI-DeepONet for all the examples presented in this work. In Figure \ref{fig:time_compare_burgers}, we present the results for the Burgers' example (1D in space and 1D in time) using networks with approximately 130,000 trainable parameters for both architectures (vanilla and separable PI-DeepONet). Meanwhile, Figure \ref{fig:time_compare_biot}, displays the comparison for consolidation based on Biot's theory (1D in space and 1D in time), with 141,802 parameters for vanilla architecture and 170,022 parameters for separable architecture. Furthermore, the application of  Sep-PI-DeepONet to the parameterized heat equation (2D in space, 1D in parameter values, and 1D in time), shown in Figure \ref{fig:heat_overview}, showcases the enhanced performance of the separable approach. The separable architecture was simulated for 2.5 hours, while the vanilla architecture approximated a completion time of 289.35 hours considering an equal number of epochs for both architectures. All computations were performed on an \textit{NVIDIA A40} graphics processing unit~(GPU) architecture. The code was written in \textit{Python} around the packages \textit{JAX}~\cite{jax2018github} and \textit{Flax}~\cite{flax2020github} as well as the \textit{Deepmind Ecosystem}~\cite{deepmind2020jax}.

\subsection{Burgers' equation}
\label{subsec:burgers}

In our first example, we consider the viscous Burgers' equation. This example has been used in \cite{Wang2021} as a benchmark example in the introductory paper of PI-DeepONet. The PDE reads as:
\begin{align}
    \frac{\partial u}{\partial t}(x, t)+u\frac{\partial u}{\partial x}(x, t)-\nu \frac{\partial^2 u}{\partial x^2}(x, t)=0, \quad\forall \; (x,t)\in[0,1]\times [0,1],
\end{align}
where $x$ and $t$ denote the spatiotemporal coordinates, $\nu=0.01$ is the kinematic viscosity, and $s$ is the fluid velocity. The periodic boundary conditions and the initial conditions are written as:
\begin{align}
    u(0,t)&=u(1,t), \quad\forall \; t\in[0,1],\\
    \frac{\partial u}{\partial x}(0, t) &= \frac{\partial u}{\partial x}(1, t), \quad\forall \; t\in[0,1],\\
    u(x,0)&=s(x), \quad\forall \; x\in[0,1],
\end{align}
where the initial condition $s(x)$ is generated from a Gaussian random process. Our goal is to learn the nonlinear solution operator, $\mathcal G_{\bm{\theta}}$ that maps the initial condition, $s(x)$ to the full spatiotemporal solution, $u(x,t)$ of the Burgers’ equation. The vanilla PI-DeepONet implementation described in \cite{Wang2021} served as our benchmark. We utilized their code to evaluate the error and computational efficiency of the vanilla framework. This approach allowed us to establish a baseline for performance comparison and accuracy verification. We sampled 2,000 initial conditions $s(x)$ from a Gaussian process with spectral density $S(k) = \sigma^2 (\tau^2 + (2\pi k)^2)^{-\gamma}$, where $\sigma = 25$, $\tau = 5$, and $\gamma = 4$, with $s(x)$ being periodic on $x \in [0,1]$. Using the inverse Fourier transform, the kernel is expressed as $K(\bm{x}, \bm{x}') = \int_{-\infty}^{\infty} S(k)\exp(2\pi i k (\bm{x} - \bm{x}')) \,\text{d}k$. The spectral discretization was evaluated with 2,048 Fourier modes in spectral form using the \textit{chebfun} package~\cite{chebfun}.

The dataset of 2,000 initial conditions was evenly split into training and testing sets ($N_{\text{train}} = N_{\text{test}} = 1,000$). The training utilized only the initial conditions, discretized uniformly at $N_{\text{IC}}=101$ points as input to the branch network, without using generated numerical solutions. Boundary conditions were evaluated at $N_{\text{BC}} = 200$ locations (100 equidistant points on each boundary, $x=0$ and $x=1$). For evaluating the PDE solution, each coordinate axis is discretized with 50 points. The vanilla architecture considers a uniform lattice grid of $N_{\text{F}}=2,500$ coordinate pairs $(x,t)$ employing a single trunk network, while the Sep-PI-DeepONet used two separate trunk networks 50 locations each for spatial and temporal domains) to evaluate the PDE residual. The training ran for 50,000 epochs using the Adam optimizer \cite{Adam2014} with a batch size of 100,000. The loss function combined PDE residual loss (2,500 collocation points), boundary condition loss (200 points), and initial condition loss (100 points, weighted by $\lambda_{\text{IC}}=20$). The initial learning rate of $1 \times 10^{-3}$ decayed exponentially at 0.95 every 1,000 epochs. All networks used \textit{Tanh} activation functions. To compare architectures, we conducted four experiments: one with the vanilla framework and three with the separable framework, varying hyperparameters $p$ and $r$. Table \ref{tab:burgers_result} reports the relative $\mathcal{L}^2$ error, total trainable parameters, and runtime for each configuration, with all models trained for 50,000 epochs. The results demonstrate that even with a comparable number of trainable parameters, both architectures achieved comparable test results, with the separable architecture showing marginally lower test accuracy. However, the separable architecture demonstrated significantly reduced training time and computational cost.

\begin{table}[!t]
    \footnotesize
    \centering
    \begin{tabular}{llllllllll}
    \hline
    &&&&& \multicolumn{3}{c}{$\mathcal{L}_2$ rel. err.} && \\
    Model & Trunk & $p$ & $r$ & Param. & Mean & Min & Max & Runtime {[}s{]} & Speedup \\ \midrule
    Vanilla & 6$\times$[100] & 100 & - & 131,701 & $5.14\rm{e}\text{-}2$ & $1.39\rm{e}\text{-}2$ & $2.86\rm{e}\text{-}1$ & 6,829.2 & - \vspace{5pt}\\
    \hdashline 
    \multirow{3}{*}{Separable (Ours)} & 6$\times$[100] & 50 & 50 & 672,151 & $6.24\rm{e}\text{-}2$ & $1.79\rm{e}\text{-}2$ & $3.39\rm{e}\text{-}1$ & 182.1  & 97,33\% \\
    & 6$\times$[100] & 20 & 20 & 244,921 & $6.04\rm{e}\text{-}2$ & $8.32\rm{e}\text{-}3$ & $3.05\rm{e}\text{-}1$ & 197.8 & 97,10\% \\
    & 6$\times$[50] & 20 & 20 & 129,221 & $6.46\rm{e}\text{-}2$ & $1.73\rm{e}\text{-}2$ & $2.94\rm{e}\text{-}1$ & 197.0  & 97,12\% \\  \bottomrule 
    \end{tabular}
    \caption{The table presents a comprehensive comparison between various Sep-PI-DeepONet configurations and a standard vanilla PI-DeepONet architecture for solving the Burgers' equation. All models have a branch of 6 layers with 100 neurons. For each model, we report the total number of trainable parameters,  mean, min, and max relative $\mathcal L_2$ error after 50,000 training epochs, total runtime, and runtime improvement, \ie, speedup (for Sep-PI-DeepONet models). The speedup metric quantifies the reduction in computational time for each Sep-PI-DeepONet configuration, expressed as a percentage relative to the vanilla PI-DeepONet's training time, which serves as the benchmark. To ensure a fair and rigorous comparison, all experiments were conducted under identical training conditions, including optimizer settings, learning rate, and hardware specifications.}
    \label{tab:burgers_result}
\end{table}

\begin{figure}[t!]
    \centering
    \includegraphics[width=1.0\textwidth]{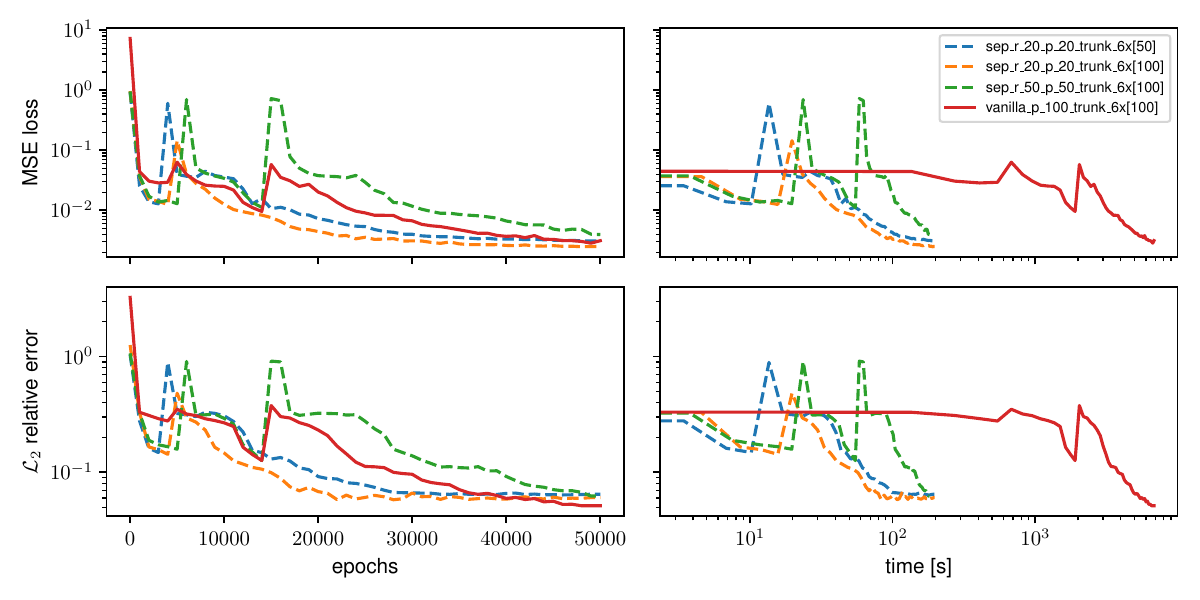}
    \caption{Comparative analysis of network architectures for the Burgers' equation. Top row: Training loss trajectories. Bottom row: Relative $\mathcal{L}_2$ error. Left: Metrics plotted against epochs. Right: Metrics plotted against computational time. While the left plots demonstrate that the convergence of all the experimental setups is similar the right plot shows that the computational time is drastically reduced in the separable architecture. All network variants listed in Table \ref{tab:burgers_result} are represented.}
    \label{fig:burgers_loss_error}
\end{figure}

\begin{figure}[h!]
    \centering
    \includegraphics[width=0.7\textwidth]{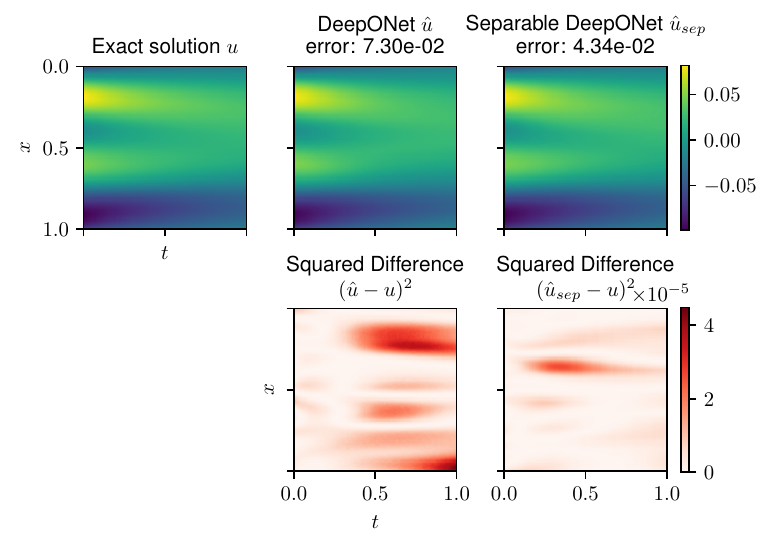}
    \caption{Burgers' equation: Performance comparison between reference vanilla PI-DeepONet (6 hidden layers, 100 neurons each, $p= 100$, resulting in 131,701 trainable parameters) and separable PI-DeepONet (6 hidden layers, 50 neurons each, $p = r = 20$, resulting in 129,221 trainable parameters) for a representative test case. Top row: Predicted solutions after 50,000 epochs. Bottom row: Squared difference between predictions and reference solution. Note the comparable accuracy as indicated by the provided relative $\mathcal{L}_2$ error for both variants despite the separable architecture's reduced complexity.}
    \label{fig:burgers_results}
\end{figure}

The Sep-PI-DeepONet with parameterization equivalent to its vanilla counterpart required only approximately 2.5\% of the reference runtime (see Table \ref{tab:burgers_result}). Here, reference runtime denotes the computational time needed by vanilla PI-DeepONet to achieve similar accuracy. Figure \ref{fig:burgers_loss_error} illustrates the loss and error curves for all variants over epochs and time, while Figure \ref{fig:burgers_results} compares the results on a test example for separable and vanilla architectures with comparable parameter counts. We observed some inconsistency between parameter count and runtime for the Sep-PI-DeepONet parametric study. For instance, the model with 672,151 parameters had a total runtime of 182 seconds, compared to approximately 197 seconds for models with 244,921 and 129,221 parameters. This inconsistency indicates potential for further optimization through improved data handling on the GPU.

\begin{figure}[h!]
    \centering
    \includegraphics[width=\textwidth]{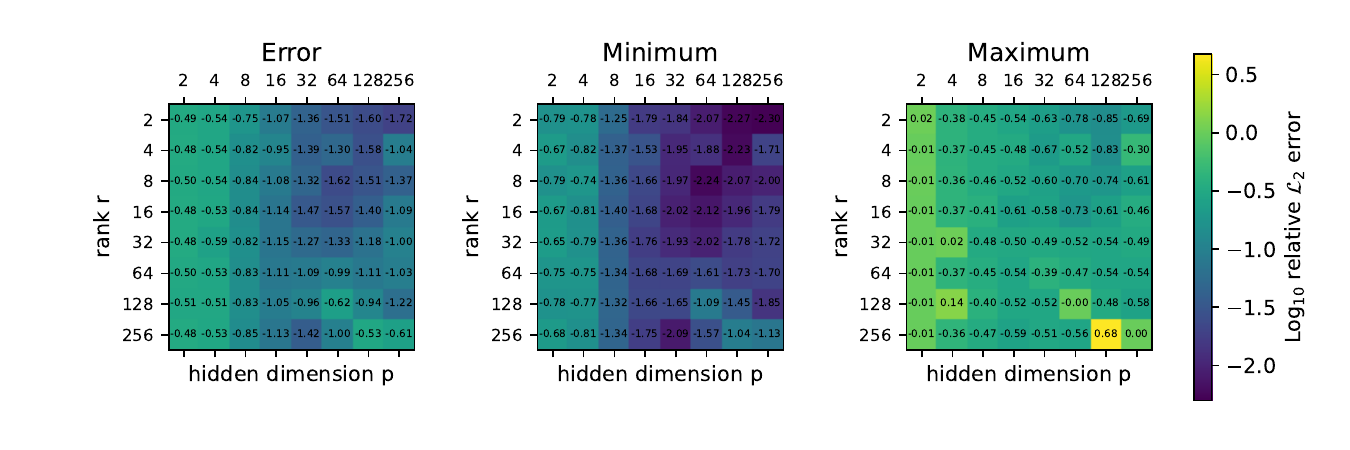}
    \caption{Influence of combinations of hidden dimension $p$ and tensor rank $r$ for Burgers' equation using Sep-PI-DeepONet with 6 trunk layers of 100 neurons. We tracked the iteration with the lowest achieved mean relative $\mathcal{L}_2$ error for each combination over training for 50,000 epochs. Additionally shown are the best and worst test examples for the respective epoch.}
    \label{fig:study_p_r}
\end{figure}

Lastly, the influence of hidden dimension $p$ and tensor rank $r$ was studied at the example of Burgers' equation. For this, a Sep-PI-DeepONet with 6 trunk layers of 100 neurons and the same settings as before was used but with all combinations for $p, r~\in~\{2, 4, 8, 16, 32, 64, 128, 256\}$. The result of this study is depicted in figure \ref{fig:study_p_r}. The smallest mean relative $\mathcal{L}_2$ error was traced through the iterations using steps of 500 throughout the training process of 50,000 epochs. Both the maximum and minimum next to the mean overall test examples are depicted. This study reveals that, with the specified hyperparameters, optimal results are achieved through combinations that feature large hidden dimension values paired with small tensor rank values. However, it is important to highlight that particularly large values of parameters $p$ and $r$ tend to lead to trivial solutions. This suggests a need for further exploration to better understand the relationship between these parameters and the impact of the training process. Consequently, we were unable to establish a universal prediction, which led us to categorize this as a classic hyperparameter issue.

\subsection{Consolidation using Biot's theory}
\label{subsec:consolidation}

Biot's theory of consolidation describes the interaction between a fluid-saturated solid body and its fluid under load~\cite{biot1941general}. This quasi-static, transient process provides insights into displacement, $u$, and fluid or pore pressure, $\Tilde{p}$. While Biot's theory is a heuristic approximation not based on balance equations or thermodynamically consistent derivation~\cite{Bertrand2022}, it serves as a useful model for coupled problems and can guide more complex applications such as advection-diffusion transport in porous media~\cite{Seyedpour2023} and active biological tissue~\cite{Tautenhahn2024}.
For our study, we consider a one-dimensional column with a permeable top and impermeable bottom, subjected to a general load at the drained surface. The governing partial differential equations are:

\begin{align}
(\lambda+2\mu)\frac{\partial^2 u}{\partial z^2}(z,t)-\frac{\partial \Tilde{p}}{\partial z}(z,t)&=0,  \quad\forall(z,t)\in[0,1]\times[0,1],\\
\frac{\partial^2 u}{\partial t\partial z}(z,t) - \frac{k}{\rho g}\frac{\partial^2 \Tilde{p}}{\partial z^2}(z,t)&=0, \quad\forall(z,t)\in[0,1]\times[0,1],
\end{align}
where $\lambda$ and $\mu$ are Lamé constants, $z$ and $t$ represent space and time, respectively, $k$ is Darcy's permeability, $\rho$ is fluid density, and $g$ is gravitational acceleration. All parameters are set to unity for simplicity. The initial and boundary conditions are:
\begin{align}
    u(z, 0)&=0,\\ 
    \Tilde{p}(z, 0)&=f(0),\label{eq:bc_con_p_ic}\\
    \sigma(0, t)&=-f(t),\label{eq:bc_con_load_func}\\
    \Tilde{p}(0,t)&=0,\\ 
    u(L, t)&=0,\\ 
    \frac{\partial \Tilde{p}}{\partial z}(L,t)&=0, 
\end{align}
where $f(t)$ is a general load function sampled using a Gaussian random process, and $L$ denotes the column length. We modified the original stress boundary condition (Equation~\eqref{eq:bc_con_load_func}) to a Dirichlet condition for displacement to simplify analysis and use of the reference solution. The derivation of these equations and the procedure to obtain an analytical solution is given in~\cite{Stickle2018}.

Our objective is to develop a non-linear solution operator $\mathcal G_{\bm \theta}$ that maps the loading function $f(t)$ to both displacement and pressure fields. We implement this using a DeepONet framework with a split trunk network~\cite{Lu2022, goswami2022neural}, where the $p$ neurons in the final layer of both branch and trunk networks are equally divided into two segments, each approximating the displacement and pore pressure fields with $p/2$ neurons. This splitting has been carried out for both frameworks.

\begin{figure}[hbt!]
    \centering
    \includegraphics[width=1.0\textwidth]{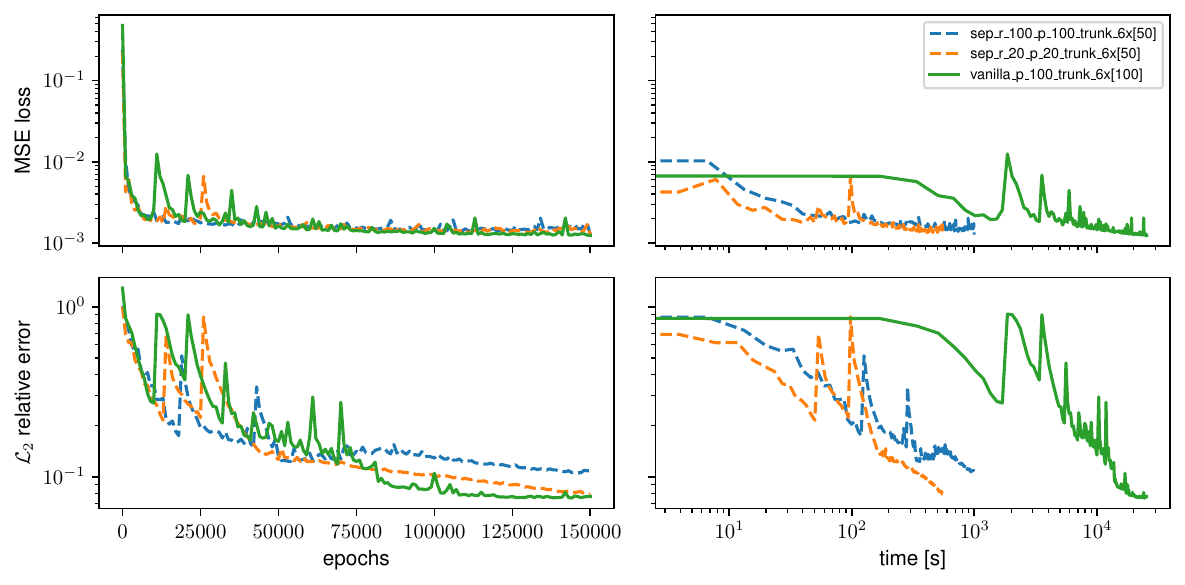}
    \caption{Convergence analysis for the consolidation problem using Biot's theory. Top row: Loss trajectories. Bottom row: Relative $\mathcal{L}_2$ error. Left column: Metrics plotted against epochs. Right column: Metrics plotted against computational time. While the left plots demonstrate that the convergence of all the experimental setups is similar the right plot shows that the computational time is drastically reduced in the separable architecture. Results are shown for all network architectures discussed in the text.}
    \label{fig:biot_loss_error}
\end{figure}

\begin{figure}[hbt!]
    \centering
    \includegraphics[width=1.0\textwidth]{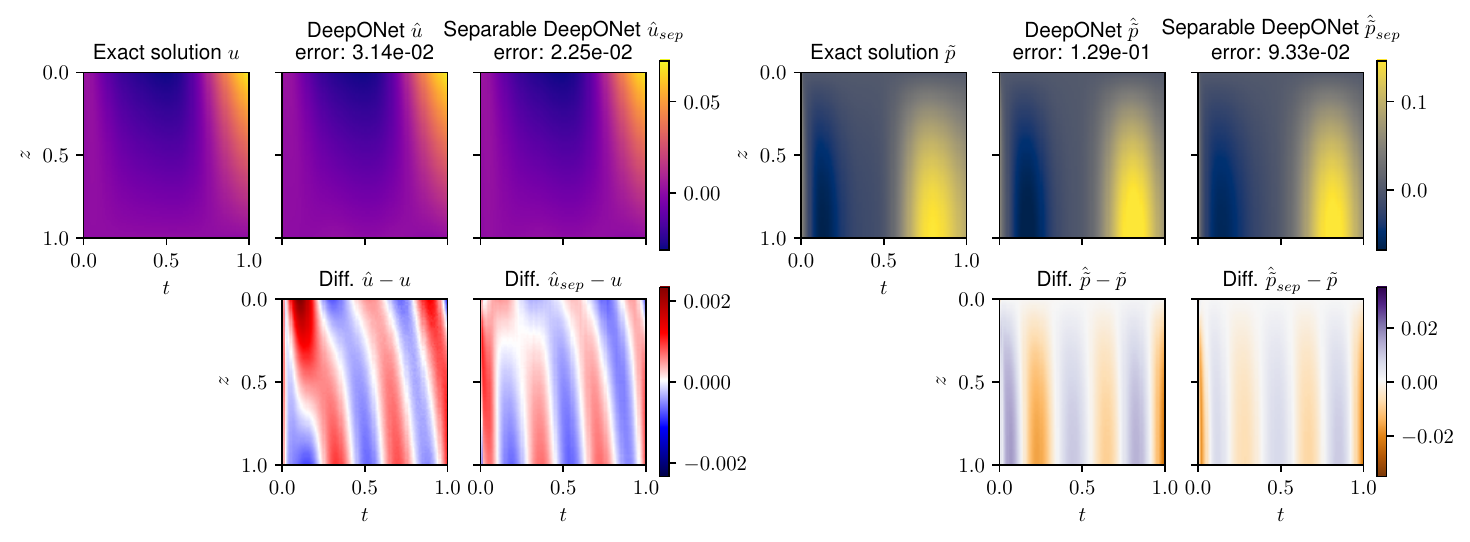}
    \caption{Comparative analysis of Biot's consolidation problem solutions: vanilla PI-DeepONet vs. Sep-PI-DeepONet. Left block: Displacement field $u$. Right block: Pressure field $\tilde{p}$. For each block: Top row: (Left) Ground truth from analytical solution; (Middle) Vanilla PI-DeepONet prediction (141,802 trainable parameters); (Right) Sep-PI-DeepONet prediction (170,022 trainable parameters). Bottom row: Difference between predicted and ground truth solutions as an absolute error to reduce the influence of large outliers in the difference}. Results are shown for a representative test sample. Both the models, Sep-PI-DeepONet and vanilla PI-DeepONet, have comparable relative $\mathcal{L}_2$ errors as indicated as errors above the respective plots.
    \label{fig:biot_results}
\end{figure}

We trained the frameworks for 150,000 epochs using the Adam optimizer~\cite{Adam2014} with exponential learning rate decay. Our dataset comprised 2,000 load functions sampled from a Gaussian process, equally split between training and testing sets ($N_{\text{train}} = N_{\text{test}} = 1000$). We validated the accuracy of testing cases against analytical step-wise solutions from~\cite{Stickle2018}. Our comparative analysis involved a vanilla DeepONet (141,802 parameters) and two Sep-DeepONet architectures (170,022 and 2,136,502 parameters). The separable architecture demonstrated significant computational efficiency improvements while achieving comparable relative $\mathcal{L}_2$ errors, despite exhibiting more rugged convergence curves (see Figures \ref{fig:biot_loss_error} and \ref{fig:biot_results}). The normal DeepONet achieved a mean relative $\mathcal{L}_2$ error of $7.66 \times 10^{-2}$ with a range from $1.20\times 10^{-2}$ to $3.40\times 10^{-1}$ per test case. The smaller Sep-DeepONet resulted in a mean relative $\mathcal{L}_2$ error of $7.83\times 10^{-2}$ ranging from $1.67\times 10^{-2}$ to $3.91\times 10^{-1}$, whereas the larger averaged in $6.17\times 10^{-2}$ with the lowest bound at $9.59\times 10^{-3}$ and the higher bound at $1.79\times 10^{-1}$ for single test cases. Training times varied significantly: the vanilla DeepONet required 25,415 seconds, while the Sep-DeepONet variants took 552 seconds (97.83\% reduction) for 170,022 parameters and 997 seconds (96.08\% reduction) for 2,136,502 parameters. These results strongly support our hypothesis that the separable architecture offers superior computational efficiency for problems amenable to the separation of variables.

\subsection{Parameterized heat equation}
\label{subsec:parametrized_heat}

As the next example, let us consider a parameterized heat equation on a two-dimensional plate. The corresponding PDE reads as:
\begin{equation}
    \frac{\partial T}{\partial t}(x,y,t) = \alpha\left(\frac{\partial^2 T}{\partial x^2}(x,y,t)+\frac{\partial^2 T}{\partial y^2}(x,y,t) \right) \quad\forall(x,y,t)\in[0,1]\times[0,1]\times[0,1],
\end{equation}
where $T$ denotes temperature, $x$, $y$, and $t$ denote the spatiotemporal coordinates, and $\alpha$ denotes the parameterized thermal diffusivity. The edges of the plate are set to temperature $T=0$ and within the domain, the temperature is set to randomly chosen constant temperature. Accordingly, the initial and boundary conditions are expressed as:
\begin{align}
    T(x,y,0) &= T_{0}, \quad 0<x<1, 0<y<1,\\
    T(0,y,t) = T(1,y,t) &= 0, \;\; \quad 0< y<1, 0\leq t\leq1,\\
    T(x,0,t) = T(x,1,t) &= 0, \;\; \quad 0< x<1, 0\leq t\leq1,
\end{align}
where $T_0$ is a predefined constant temperature. The initial temperature field $T_{0}$ is taken as input to the branch network, while the spatiotemporal coordinates and thermal diffusivity $\alpha$ are input to the respective trunk networks. The initial temperature $T_{0}$ is considered in the range $[0,1]$, while the thermal diffusivity spans several magnitudes within $[10^{-2}, 10^{0}]$. To incorporate this into our model, we used a parameter $c=\sqrt{\alpha}$ and squared this factor in the residual calculation.

The Sep-PI-DeepONet architecture consisted of 6 hidden layers with 50 neurons each in the trunk and branch networks, with hidden dimension and tensor rank set to $p=r=50$. This configuration resulted in 576,801 total parameters with 4 separable trunk networks. The model was trained for $100,000$ epochs using the Adam optimizer~\cite{Adam2014} with an initial learning rate of $1 \times 10^{-3}$ and an exponential decay rate of $0.9$ every $1,500$ step. The total training time was $9173$s. During training, $N_{\text{train}} = 25$ initial temperatures were sampled uniformly from $[0,1]$. The initial condition was evaluated on a $51 \times 51$ grid of spatial points, while boundary conditions at $x=0$, $x=1$, $y=0$, and $y=1$ were sampled with $51 \times 51$ equidistant points in the remaining spatial dimension and time. Values of $\alpha$ were sampled at 51 points between $[10^{-2}, 10^{0}]$, and the residual was evaluated at factorized coordinates with a uniform spacing of 31 points in $x$, $y$, $t$, and $\alpha$.

\begin{figure}[hbt!]
    \centering
    \includegraphics[width=1.0\textwidth]{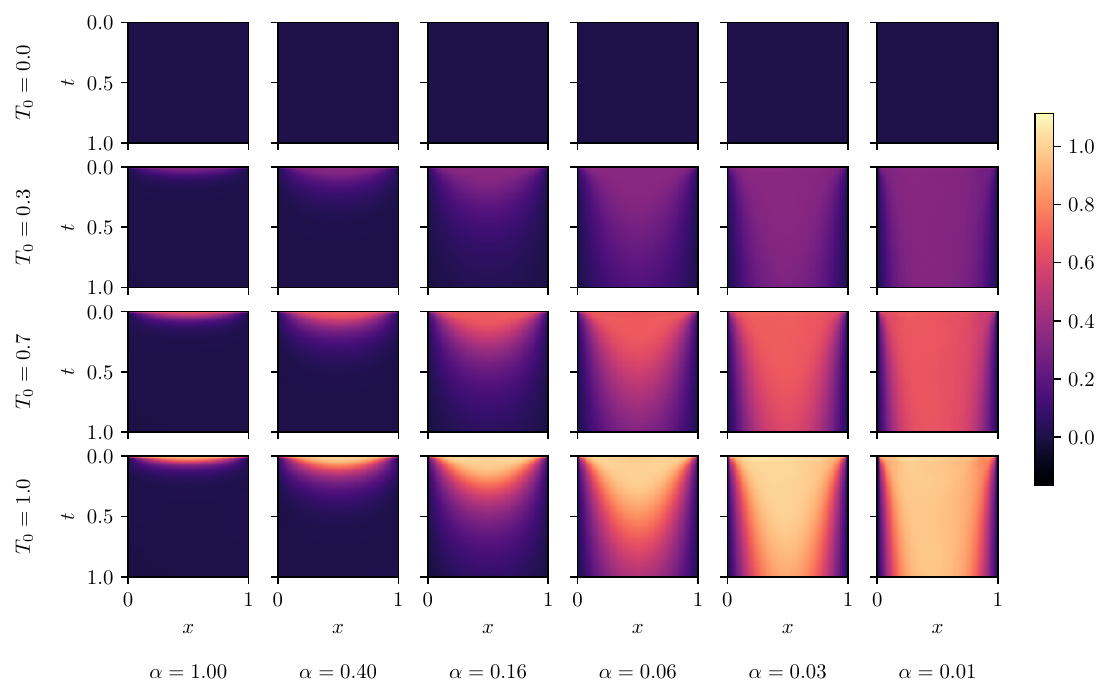}
    \caption{Predictions using the trained Sep-PI-DeepONet for the two-dimensional heat-equation along the middle plane ($y=0.5$) of the domain with several input temperatures and heat. This demonstrates the capabilities of Sep-DeepONet to make predictions over the whole training domain even for values not included during training.}
    \label{fig:heat_solved}
\end{figure}

The trained Sep-PI-DeepONet was evaluated on $N_{\text{test}} = 200$ test examples, successfully solving the heat equation for given initial temperatures and thermal diffusivity throughout the spatial domain. Solutions were compared to analytical results obtained by separation of variables and Fourier series analysis~\cite{Fourier1822}. The model achieved a final relative $\mathcal{L}_2$ error of $8.56 \times 10^{-2}$, indicating a good approximation of the analytical solution (see Figure \ref{fig:heat_solved}) with values for individual initial temperatures ranging from $3.22 \times 10^{-2}$ to $2.05 \times 10^{-1}$ for the relative $\mathcal{L}_2$ error. Notably, the Sep-PI-DeepONet demonstrated a significant advantage in training time compared to a vanilla PI-DeepONet, completing training in just over $2.5$ hours versus an estimated $289.35$ hours for the vanilla model—an empirical speedup factor exceeding $100$. This substantial improvement in training efficiency highlights the benefits of the separable architecture in overcoming the curse of dimensionality inherent in high-dimensional parametric PDE problems. The successful application of Sep-PI-DeepONet to the parameterized heat equation showcases its potential for efficiently solving complex, high-dimensional partial differential equation problems, demonstrating promise for advancing scientific computing and numerical simulation.

\begin{figure}[hbt!]
    \centering
    \begin{subfigure}[b]{0.595\textwidth}
         \centering
        \includegraphics[width=\textwidth]{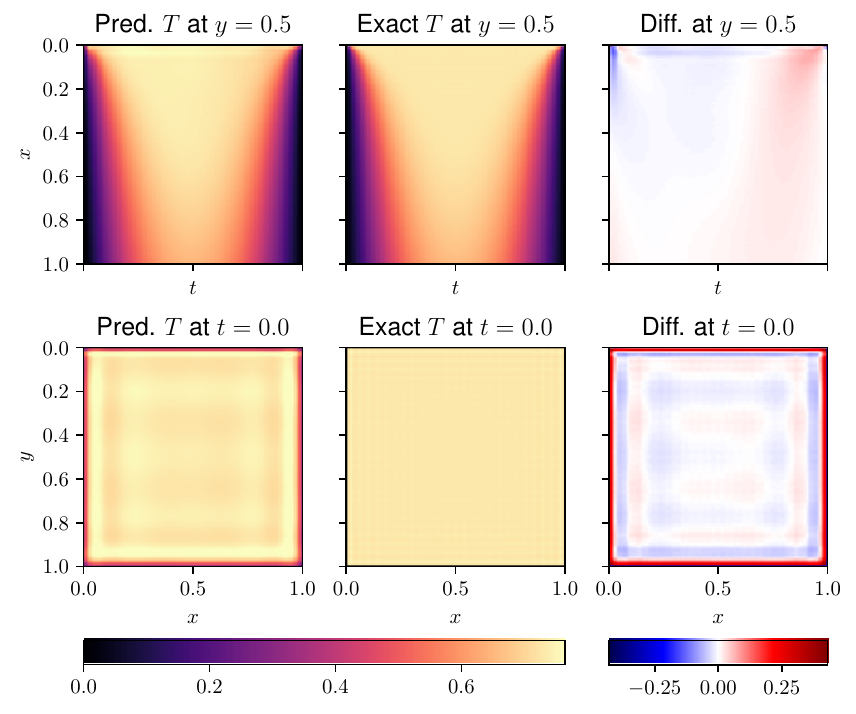}
         \caption{Prediction, analytical solution and difference at $y=0.5$}
         \label{fig:field_heat}
     \end{subfigure}
     \hfill
     \begin{subfigure}[b]{0.395\textwidth}
         \centering
         \includegraphics[width=\textwidth]{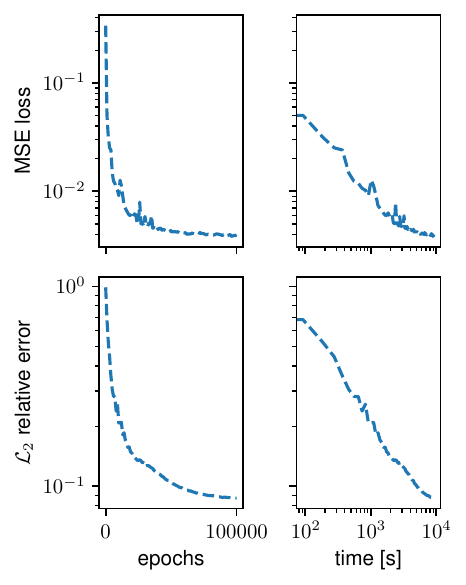}
         \caption{Loss and error plots}
         \label{fig:loss_error_heat}
     \end{subfigure}
    \caption{Comparison of the Sep-PI-DeepONet predictions and the analytical solution for the heat equation with $\alpha=0.159$ and $T_0=0.2$ (Figure \ref{fig:field_heat}). The plot shows the temperature field along the central plane at $y=0.5$ and the initial time step $t=0$. While the predictions capture the overall trend, the largest discrepancies are observed near the boundaries, leading to a relative $\mathcal{L}_2$ error of $8.56 \times 10^{-2}$ for this test case. The training loss and test error evolution during the 100,000 training epochs are depicted in (b), demonstrating the model's convergence.}
    \label{fig:heat_result}
\end{figure}

\subsection{Parameterized Poisson's equation with random field source}
\label{subsec:parametrized_poisson}
\begin{figure}[hbt!]
    \centering
    \includegraphics[width=1.0\textwidth]{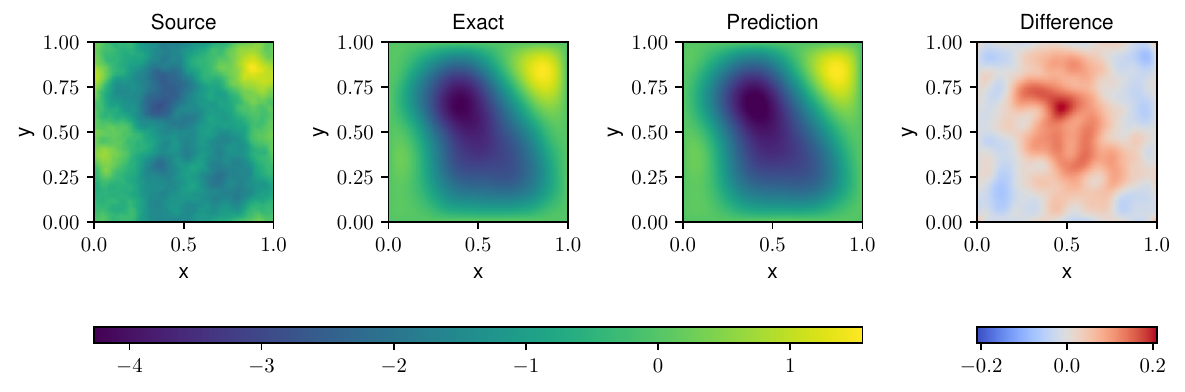}
    \caption{A representative result of the Sep-CNN-DeepONet as described is shown here. As the diffusivity effectively scales the result, we only depict one result for $k=0.01$ and a relative $\mathcal{L}_2$ error of $3.46 \times 10^{-2}$ alongside the source field on the left and the absolute difference on the right.}
    \label{fig:poisson_solved}
\end{figure}

The final example demonstrates the method's versatility by solving Poisson's equation with a random field source in two dimensions. While this problem was previously investigated in~\cite{Koric2023}, where classical and PI-DeepONets were compared, we extend the approach by implementing a separable framework in the trunk network and incorporating a convolutional neural network (CNN) in the branch network, establishing a PI-Sep-CNN-DeepONet. The governing equation is:
\begin{equation}
u(x,y) = k\left(\frac{\partial^2 S}{\partial x^2}(x,y)+\frac{\partial^2 S}{\partial y^2}(x,y) \right) \quad \forall(x,y)\in[0,1]\times[0,1],
\end{equation}
where $S$ represents the field variable, $(x,y)$ denotes spatial coordinates, $k$ is the diffusion coefficient, and $u$ represents the source term. The source term is sampled as a normalized Gaussian random field with a spatial correlation following $P(k) \sim {1}/{{\vert k\vert}^{\alpha/2}}$, where $\alpha=4$~\cite{bsciolla2017}. The problem is subject to homogeneous Dirichlet boundary conditions:
\begin{equation}
S(x,0) = S(x, 1) = S(0, y) = S(1, y) = 0.
\end{equation}
The input function was sampled on a uniform grid of $51\times 51$ points as branch input. The branch CNN consisted of 3 convolution layers with 16, 32, and 64 filters each followed by an average pooling layer each. Output is then aggregated by two dense layers with 400 neurons each and \textit{ReLU} activation.  Input to the trunk network are the two spatial dimensions as well as the diffusion coefficients. The former was sampled with the same 51 points per dimension as the input, while the diffusion coefficient was sampled logarithmically from $10^{-3}$ to $10^{1}$ at 21 points. Each trunk network consisted of 6 hidden layers with 50 neurons. The hidden dimension was set to $p=400$ following the reference work and tensor rank to $r=80$. The complete network architecture comprises 6,187,767 parameters. For training and validation, we generated 5,000 random fields, splitting them equally between training and testing sets. The test cases evaluated diffusivity values of $10^{-2}$, $10^{-1}$, and $10^{0}$ on the $51\times 51$ grid. The training protocol involved 60,000 epochs with an exponential learning rate decay factor of 0.8 applied every 5,000 steps, starting from an initial learning rate of $1 \times 10^{-3}$. A notable characteristic of the problem is that the diffusivities exhibit an inverse relationship with the sought solution, such that larger values of $k$ result in smaller field values. Since the absolute error is optimized across all cases, while we track the relative error, a larger relative error is inherently observed with higher diffusivities. Nevertheless, we obtain an average relative $\mathcal{L}_2$ error of $4.73 \times 10^{-2}$ for all test cases after a training time of 2,132 seconds, with the maximum error occurring at $2.19 \times 10^{-2}$, and the minimum at $1.79 \times 10^{-1}$. The result for a single random field and diffusivity is shown in figure \ref{fig:poisson_solved}.

\section{Discussion and Summary}
\label{sec:conclusion}

The domain of physics-informed DeepONet has remained largely unexplored due to the resource-intensive nature of its training process. This work introduces a separable architecture for physics-informed DeepONet, inspired by the separation of variables technique. The efficiency of our framework stems from two key innovations: leveraging factorized inputs through independent trunk networks for each coordinate axis and employing forward mode automatic differentiation. This approach significantly reduces computational complexity and training time, enabling the application of physics-informed DeepONet to large-scale problems, particularly those amenable to separation of variables techniques.
In this work, we demonstrated through three examples that the separable architecture simulates solutions with two orders of magnitude lower computational time while achieving comparable accuracy. Furthermore, our application to the heat equation shows that Sep-PI-DeepONet can effectively tackle problems where the vanilla version is computationally intractable, thus showcasing significant advantages over vanilla PI-DeepONet. This significant reduction in computational requirements, coupled with maintained solution quality, highlights the potential of Sep-PI-DeepONet to expand the scope and scale of problems addressable by physics-informed neural networks.

While Sep-PI-DeepONet demonstrates promising advantages across various applications, particularly for parameterized problems, the method faces several inherent limitations in its current configuration. The fundamental challenge lies in the use of factorized inputs: although this approach enables efficient handling of high-dimensional data, it requires the underlying problem to be amenable to factorization in both parameters and coordinates. These limitations can manifest through multiple mechanisms, including the nature of the governing equations, the problem characteristics, and the chosen coordinate system. Two significant challenges arise: (1) the handling of irregular geometries, and (2) the treatment of coefficient combinations that are incompatible with the problem domain. The latter can be addressed through sophisticated preprocessing techniques, such as training with valid combination batches or implementing selective loss calculations during postprocessing. However, these solutions may increase memory requirements without necessarily incurring additional computational costs. The constraint of regular geometries becomes particularly problematic when dealing with real-world measurement data and domains that deviate from the required grid-based structure. Potential mitigation strategies include incorporating problem-specific, tailor-made architectures within the trunk network to introduce factorizable sampling, leveraging the framework's inherent parallelization capabilities, and extending the separable approach to branch network inputs, which could enable a more versatile operator learning framework capable of handling combinations of input functions. The development of a modified version of this framework for PDEs which inherently cannot be solved using the separation of variables technique will be an important future direction. Addressing these challenges could enhance Sep-PI-DeepONet's generalization ability across a wide range of high-dimensional parametric PDE problems. 

Although the Sep-DeepONet and Sep-PI-DeepONet architectures have significantly reduced computational demands during forward passes and automatic differentiation compared to their conventional counterparts, memory consumption remains a critical bottleneck even on state-of-the-art high-performance computing systems and GPU architectures. This limitation becomes particularly challenging when incorporating additional parameterizations, such as variable geometries or higher-dimensional sampling spaces, especially if the separable framework is extended to branch network input.

While this study primarily focused on unstacked configurations, both stacked and unstacked DeepONet architectures have demonstrated effectiveness in various settings~\cite{lu2021learning,Lu2022}. This opens avenues for further research, including strategies for handling multiple inputs and outputs~\cite{jin2022mionet,goswami2022neural}, multi-fidelity data~\cite{howard2023multifidelity}, and multi-modal data structures. Although we employed classical feed-forward neural networks in this study, the Sep-DeepONet concept applies to other architectures such as convolutional, recurrent, and residual networks~\cite{Wang2021MLP}. Furthermore, Sep-PI-DeepONet can be modified similarly to standard PI-DeepONet, allowing for the incorporation of advanced techniques like causality-respecting loss functions~\cite{Wang2024}, variational formulations~\cite{goswami2022physics}, and hard constraint boundary conditions~\cite{Lu2022}. These potential extensions and modifications highlight the versatility and adaptability of the Sep-DeepONet framework, suggesting promising directions for future research and applications in computational physics and scientific machine learning.

The enhanced efficiency and scalability of Sep-PI-DeepONet not only broadens its applicability but also facilitates the possibility of more comprehensive hyperparameter tuning. This, in turn, will lead to improved generalization capabilities, potentially expanding the range of complex physical systems that can be accurately modeled and simulated using neural network approaches. As such, Sep-PI-DeepONet represents a significant step forward in making physics-informed machine learning more practical and accessible for challenging scientific and engineering problems.

\bibliographystyle{plain}
\bibliography{scibib}

\begin{thebibliography}{10}

\bibitem{anandkumar2020neural}
Anima Anandkumar, Kamyar Azizzadenesheli, Kaushik Bhattacharya, Nikola Kovachki, Zongyi Li, Burigede Liu, and Andrew Stuart.
\newblock {Neural Operator: Graph Kernel Network for Partial Differential Equations}.
\newblock In {\em ICLR 2020 Workshop on Integration of Deep Neural Models and Differential Equations}, 2020.

\bibitem{baydin2018}
Atilim~Gunes Baydin, Barak~A. Pearlmutter, Alexey~Andreyevich Radul, and Jeffrey~Mark Siskind.
\newblock Automatic differentiation in machine learning: a survey.
\newblock {\em Journal of Machine Learning Research}, 18(153):1--43, 2018.

\bibitem{Bertrand2022}
Fleurianne Bertrand, Maximilian Brodbeck, and Tim Ricken.
\newblock On robust discretization methods for poroelastic problems: Numerical examples and counter-examples.
\newblock {\em Examples and Counterexamples}, 2:100087, November 2022.

\bibitem{billaud2014tensor}
Marie Billaud-Friess, Anthony Nouy, and Olivier Zahm.
\newblock A tensor approximation method based on ideal minimal residual formulations for the solution of high-dimensional problems.
\newblock {\em ESAIM: Mathematical Modelling and Numerical Analysis}, 48(6):1777--1806, 2014.

\bibitem{biot1941general}
Maurice~A Biot.
\newblock {General Theory of Three‐Dimensional Consolidation}.
\newblock {\em Journal of Applied Physics}, 12(2):155--164, 1941.

\bibitem{jax2018github}
James Bradbury, Roy Frostig, Peter Hawkins, Matthew~James Johnson, Chris Leary, Dougal Maclaurin, George Necula, Adam Paszke, Jake Vander{P}las, Skye Wanderman-{M}ilne, and Qiao Zhang.
\newblock {JAX}: composable transformations of {P}ython+{N}um{P}y programs, 2018.

\bibitem{cao2023lno}
Qianying Cao, Somdatta Goswami, and George~Em Karniadakis.
\newblock {LNO: Laplace Neural Operator for Solving Differential Equations}.
\newblock {\em arXiv preprint arXiv:2303.10528}, 2023.

\bibitem{chen1995universal}
Tianping Chen and Hong Chen.
\newblock {Universal approximation to nonlinear operators by neural networks with arbitrary activation functions and its application to dynamical systems}.
\newblock {\em IEEE Transactions on Neural Networks}, 6(4):911--917, 1995.

\bibitem{chinesta2013proper}
Francisco Chinesta, Roland Keunings, and Adrien Leygue.
\newblock {\em The proper generalized decomposition for advanced numerical simulations: a primer}.
\newblock Springer Science \& Business Media, 2013.

\bibitem{cho2023separable}
Junwoo Cho, Seungtae Nam, Hyunmo Yang, Seok-Bae Yun, Youngjoon Hong, and Eunbyung Park.
\newblock Separable physics-informed neural networks.
\newblock {\em Advances in Neural Information Processing Systems}, 2023.

\bibitem{de2008tensor}
Vin De~Silva and Lek-Heng Lim.
\newblock Tensor rank and the ill-posedness of the best low-rank approximation problem.
\newblock {\em SIAM Journal on Matrix Analysis and Applications}, 30(3):1084--1127, 2008.

\bibitem{deepmind2020jax}
DeepMind, Igor Babuschkin, Kate Baumli, Alison Bell, Surya Bhupatiraju, Jake Bruce, Peter Buchlovsky, David Budden, Trevor Cai, Aidan Clark, Ivo Danihelka, Antoine Dedieu, Claudio Fantacci, Jonathan Godwin, Chris Jones, Ross Hemsley, Tom Hennigan, Matteo Hessel, Shaobo Hou, Steven Kapturowski, Thomas Keck, Iurii Kemaev, Michael King, Markus Kunesch, Lena Martens, Hamza Merzic, Vladimir Mikulik, Tamara Norman, George Papamakarios, John Quan, Roman Ring, Francisco Ruiz, Alvaro Sanchez, Laurent Sartran, Rosalia Schneider, Eren Sezener, Stephen Spencer, Srivatsan Srinivasan, Milo\v{s} Stanojevi\'{c}, Wojciech Stokowiec, Luyu Wang, Guangyao Zhou, and Fabio Viola.
\newblock The {D}eep{M}ind {JAX} {E}cosystem, 2020.

\bibitem{chebfun}
Tobin~A. Driscoll, Nicholas Hale, and Lloyd~N. Trefethen.
\newblock Chebfun guide, 2014.

\bibitem{Fourier1822}
Jean Baptiste~Joseph Fourier.
\newblock {\em Théorie Analytique de la Chaleur}.
\newblock Firmin Didot, Paris, 1822.

\bibitem{goswami2022neural}
Somdatta Goswami, David~S Li, Bruno~V Rego, Marcos Latorre, Jay~D Humphrey, and George~Em Karniadakis.
\newblock {Neural operator learning of heterogeneous mechanobiological insults contributing to aortic aneurysms}.
\newblock {\em Journal of the Royal Society Interface}, 19(193):20220410, 2022.

\bibitem{goswami2022physics}
Somdatta Goswami, Minglang Yin, Yue Yu, and George~Em Karniadakis.
\newblock {A physics-informed variational DeepONet for predicting crack path in quasi-brittle materials}.
\newblock {\em Computer Methods in Applied Mechanics and Engineering}, 391:114587, 2022.

\bibitem{He2024}
Junyan He, Seid Koric, Diab Abueidda, Ali Najafi, and Iwona Jasiuk.
\newblock Geom-deeponet: A point-cloud-based deep operator network for field predictions on 3d parameterized geometries.
\newblock {\em Computer Methods in Applied Mechanics and Engineering}, 429:117130, September 2024.

\bibitem{He2024b}
Junyan He, Shashank Kushwaha, Jaewan Park, Seid Koric, Diab Abueidda, and Iwona Jasiuk.
\newblock Sequential deep operator networks (s-deeponet) for predicting full-field solutions under time-dependent loads.
\newblock {\em Engineering Applications of Artificial Intelligence}, 127:107258, January 2024.

\bibitem{flax2020github}
Jonathan Heek, Anselm Levskaya, Avital Oliver, Marvin Ritter, Bertrand Rondepierre, Andreas Steiner, and Marc van {Z}ee.
\newblock {F}lax: A neural network library and ecosystem for {JAX}, 2023.

\bibitem{howard2023multifidelity}
Amanda~A Howard, Mauro Perego, George~Em Karniadakis, and Panos Stinis.
\newblock {Multifidelity deep operator networks for data-driven and physics-informed problems}.
\newblock {\em Journal of Computational Physics}, 493:112462, 2023.

\bibitem{jin2022mionet}
Pengzhan Jin, Shuai Meng, and Lu~Lu.
\newblock {MIONet: Learning multiple-input operators via tensor product}.
\newblock {\em SIAM Journal on Scientific Computing}, 44(6):A3490--A3514, 2022.

\bibitem{Adam2014}
Diederik~P. Kingma and Jimmy Ba.
\newblock {Adam: A Method for Stochastic Optimization}, 2014.

\bibitem{Kobayashi2024}
Kazuma Kobayashi, James Daniell, and Syed~Bahauddin Alam.
\newblock Improved generalization with deep neural operators for engineering systems: Path towards digital twin.
\newblock {\em Engineering Applications of Artificial Intelligence}, 131:107844, May 2024.

\bibitem{kontolati2023influence}
Katiana Kontolati, Somdatta Goswami, Michael~D Shields, and George~Em Karniadakis.
\newblock On the influence of over-parameterization in manifold based surrogates and deep neural operators.
\newblock {\em Journal of Computational Physics}, 479:112008, 2023.

\bibitem{Koric2023}
Seid Koric and Diab~W. Abueidda.
\newblock Data-driven and physics-informed deep learning operators for solution of heat conduction equation with parametric heat source.
\newblock {\em International Journal of Heat and Mass Transfer}, 203:123809, April 2023.

\bibitem{Kushwaha2024}
Shashank Kushwaha, Jaewan Park, Seid Koric, Junyan He, Iwona Jasiuk, and Diab Abueidda.
\newblock Advanced deep operator networks to predict multiphysics solution fields in materials processing and additive manufacturing.
\newblock {\em Additive Manufacturing}, 88:104266, May 2024.

\bibitem{Leng2024}
Kuangdai Leng, Mallikarjun Shankar, and Jeyan Thiyagalingam.
\newblock Zero coordinate shift: Whetted automatic differentiation for physics-informed operator learning.
\newblock {\em Journal of Computational Physics}, 505:112904, May 2024.

\bibitem{Li2023}
Wei Li, Martin~Z. Bazant, and Juner Zhu.
\newblock Phase-field deeponet: Physics-informed deep operator neural network for fast simulations of pattern formation governed by gradient flows of free-energy functionals, 2023.

\bibitem{li2021fourier}
Zongyi Li, Nikola~Borislavov Kovachki, Kamyar Azizzadenesheli, Burigede liu, Kaushik Bhattacharya, Andrew Stuart, and Anima Anandkumar.
\newblock {Fourier Neural Operator for Parametric Partial Differential Equations}.
\newblock In {\em In Proceedings of the International Conference on Learning Representations}, 2021.

\bibitem{lu2021learning}
Lu~Lu, Pengzhan Jin, Guofei Pang, Zhongqiang Zhang, and George~Em Karniadakis.
\newblock {Learning nonlinear operators via DeepONet based on the universal approximation theorem of operators}.
\newblock {\em Nature machine intelligence}, 3(3):218--229, 2021.

\bibitem{Lu2022}
Lu~Lu, Xuhui Meng, Shengze Cai, Zhiping Mao, Somdatta Goswami, Zhongqiang Zhang, and George~Em Karniadakis.
\newblock {A comprehensive and fair comparison of two neural operators (with practical extensions) based on FAIR data}.
\newblock {\em Computer Methods in Applied Mechanics and Engineering}, 393:114778, April 2022.

\bibitem{bsciolla2017}
Bruno Sciolla.
\newblock {Generator of 2D Gaussian Random Fields}.

\bibitem{Seyedpour2023}
S.~M. Seyedpour, A.~Thom, and T.~Ricken.
\newblock Simulation of contaminant transport through the vadose zone: A continuum mechanical approach within the framework of the extended theory of porous media (etpm).
\newblock {\em Water}, 15(2):343, January 2023.

\bibitem{Stickle2018}
M.~M. Stickle and M.~Pastor.
\newblock {A practical analytical solution for one-dimensional consolidation}.
\newblock {\em Géotechnique}, 68(9):786–793, September 2018.

\bibitem{Tautenhahn2024}
Hans‐Michael Tautenhahn, Tim Ricken, Uta Dahmen, Luis Mandl, Laura B\"{u}tow, Steffen Gerh\"{a}usser, Lena Lambers, Xinpei Chen, Elina Lehmann, Olaf Dirsch, and Matthias K\"{o}nig.
\newblock Simliva–modeling ischemia‐reperfusion injury in the liver: A first step towards a clinical decision support tool.
\newblock {\em GAMM-Mitteilungen}, 47(2), January 2024.

\bibitem{tripura2023wavelet}
Tapas Tripura and Souvik Chakraborty.
\newblock {Wavelet Neural Operator for solving parametric partial differential equations in computational mechanics problems}.
\newblock {\em Computer Methods in Applied Mechanics and Engineering}, 404:115783, 2023.

\bibitem{vella2022pgd}
Cl{\'e}ment Vella and Serge Prudhomme.
\newblock Pgd reduced-order modeling for structural dynamics applications.
\newblock {\em Computer Methods in Applied Mechanics and Engineering}, 402:115736, 2022.

\bibitem{Wang2024}
Sifan Wang, Shyam Sankaran, and Paris Perdikaris.
\newblock Respecting causality for training physics-informed neural networks.
\newblock {\em Computer Methods in Applied Mechanics and Engineering}, 421:116813, March 2024.

\bibitem{Wang2021MLP}
Sifan Wang, Yujun Teng, and Paris Perdikaris.
\newblock Understanding and mitigating gradient flow pathologies in physics-informed neural networks.
\newblock {\em SIAM Journal on Scientific Computing}, 43(5):A3055–A3081, January 2021.

\bibitem{Wang2021}
Sifan Wang, Hanwen Wang, and Paris Perdikaris.
\newblock {Learning the solution operator of parametric partial differential equations with physics-informed DeepONets}.
\newblock {\em Science Advances}, 7(40), October 2021.

\bibitem{yu2024separable}
Xinling Yu, Sean Hooten, Ziyue Liu, Yequan Zhao, Marco Fiorentino, Thomas Van~Vaerenbergh, and Zheng Zhang.
\newblock Separable operator networks.
\newblock {\em arXiv preprint arXiv:2407.11253}, 2024.

\bibitem{Zhang2024}
Jiahao Zhang, Shiheng Zhang, Jie Shen, and Guang Lin.
\newblock Energy-dissipative evolutionary deep operator neural networks.
\newblock {\em Journal of Computational Physics}, 498:112638, February 2024.

\end{thebibliography}

\section*{Funding}

L.M. and T.R. were supported by Deutsche Forschungsgemeinschaft~(DFG, German Research Foundation) by grant number 465194077~(Priority Programme SPP 2311,~Project SimLivA). L.M. and L.L. are supported by the Add-on Fellowship of the Joachim Herz Foundation. L.L. and T.R. were supported by the Federal Ministry of Education and Research (BMBF, Germany) within ATLAS by grant number 031L0304A and by Deutsche Forschungsgemeinschaft (DFG, German Research Foundation) under Germany's Excellence Strategy – EXC 2075 – 390740016. T.R. thanks the Deutsche Forschungsgemeinschaft (DFG, German Research Foundation) for support via the projects ``Hybrid MOR" by grant number 504766766 and FOR 5151 QuaLiPerF (Quantifying Liver Perfusion-Function Relationship in Complex Resection - A Systems Medicine Approach)" by grant number 436883643. S.G. is supported by the U.S. Department of Energy, Office of Science, Office of Advanced Scientific Computing Research grant under Award Number DE-SC0024162.

\section*{Data and code availability}
All data needed to evaluate the conclusions in the paper are presented in the paper and/or the Supplementary Materials. All code and data generation scripts accompanying this manuscript are available at \url{https://github.com/lmandl/separable-PI-DeepONet}.

\section*{Author contributions}
\noindent 
Conceptualization: L.M., S.G., L.L., T.R. \\
Investigation: L.M., S.G., L.L. \\
Visualization: L.M., S.G., L.L. \\
Supervision: S.G., T.R. \\
Writing—original draft: L.M., S.G. \\
Writing—review \& editing: L.M., S.G., L.L., T.R.

\newpage  
\renewcommand{\thetable}{S\arabic{table}}  
\renewcommand{\thefigure}{S\arabic{figure}}
\renewcommand{\thealgocf}{S\arabic{algocf}}
\makeatother
\setcounter{figure}{0}
\setcounter{table}{0}
\setcounter{section}{0}
\setcounter{algocf}{0}
\setcounter{page}{1}

{\Large{\textbf{\center{Supplementary Materials}}}}

\renewcommand{\thesection}{S\arabic{section}}

\section{Theoretical details of DeepONet}
\label{sec:theoretical-details} 

Let $\Omega \subset \mathbb{R}^D$ be a bounded open set and $\mathcal{X}=\mathcal{X}(\Omega; \mathbb{R}^{d_x})$ and $\mathcal{Y}=\mathcal{Y}(\Omega;\mathbb{R}^{d_y})$ two separable Banach spaces. Furthermore, assume that $\mathcal{G}: \mathcal{X} \rightarrow \mathcal{Y}$ is a non-linear map arising from the solution of a time-dependent PDE. The objective is to approximate the nonlinear operator via the following parametric mapping
\begin{equation}
\begin{aligned}
    \mathcal{G}: \mathcal{X} \times \Theta \rightarrow \mathcal{Y} \hspace{15pt} \text{or}, \hspace{15pt} \mathcal{G}_{\theta}: \mathcal{X} \rightarrow \mathcal{Y}, \hspace{5pt} \theta \in \Theta
\end{aligned}
\end{equation}
where $\Theta$ is a finite-dimensional parameter space. The optimal parameters $\theta^*$ are learned via the training of a neural operator with backpropagation based on a dataset $\{\mathbf{x}_j, \mathbf{y}_j \}_{j=1}^N$ generated on a discretized domain $\Omega_m = \{x_1, \dots, x_m\} \subset \Omega$ where $\{x_j\}_{j=1}^m$ represent the sensor locations, thus $\mathbf{x}_{j|\Omega_m} \in \mathbb{R}^{D_x}$ and $\mathbf{y}_{j|\Omega_m} \in \mathbb{R}^{D_y}$ where $D_x= d_x \times m$ and $D_y = d_y \times m$.

The Deep Operator Network (DeepONet) \cite{lu2021learning} aims to learn operators between infinite-dimensional Banach spaces. Learning is performed in a general setting in the sense that the sensor locations $\{x_i\}_{i=1}^m$ at which the input functions are evaluated need not be equispaced, however, they need to be consistent across all input function evaluations. Instead of blindly concatenating the input data (input functions $[\mathbf{x}(x_1), \mathbf{x}(x_2), \dots, \mathbf{x}(x_m)]^T$ and locations $\zeta$)
as one input, \ie, $[\mathbf{x}(x_1), \mathbf{x}(x_2), \dots, \mathbf{x}(x_m), \zeta]^T$, DeepONet employs two subnetworks and treats the two inputs equally. Thus, DeepONet can be applied for high-dimensional problems, where the dimension of $\mathbf{x}(x_i)$ and $\zeta$ no longer match since the latter is a vector of $d$ components in total. A trunk network $\mathbf{f}(\cdot)$, takes as input $\zeta$ and outputs $[tr_1, tr_2, \ldots, tr_p]^T \in \mathbb{R}^p$ while a second network, the branch net $\mathbf{g}(\cdot)$, takes as input $[\mathbf{x}(x_1), \mathbf{x}(x_2), \dots, \mathbf{x}(x_m)]^T$ and outputs $[b_1, b_2, \ldots, b_p]^T \in \mathbb{R}^p$. Both subnetwork outputs are merged through a dot product to generate the quantity of interest. A bias $b_0 \in \mathbb{R}$ is added in the last stage to increase expressivity, \ie, $\mathcal{G}(\mathbf{x})(\zeta) \approx \sum_{i=k}^p b_k t_k + b_0$. The generalized universal approximation theorem for operators, inspired by the original theorem introduced by \cite{chen1995universal}, is presented below. The generalized theorem essentially replaces shallow networks used for the branch and trunk net in the original work with deep neural networks to gain expressivity.

\begin{theorem}[Generalized Universal Approximation Theorem for Operators.]
\label{pythagorean}
Suppose that $X$ is a Banach space, $K_1 \subset X$, $K_2 \subset \mathbb{R}^d$ are two compact sets in $X$ and $\mathbb{R}^d$, respectively, $V$ is a compact set in $C(K_1)$. Assume that: $\mathcal{G}: V \rightarrow C(K_2)$ is a nonlinear continuous operator. Then, for any $\epsilon > 0$, there exist positive integers $m, p$, continuous vector functions $\mathbf{g}: \mathbb{R}^m \rightarrow \mathbb{R}^p$, $\mathbf{f}: \mathbb{R}^d \rightarrow \mathbb{R}^p$, and $x_1, x_2, \dots , x_m \in K_1$ such that   
\[  \Bigg\lvert \mathcal{G}(\mathbf{x})(\zeta) - \langle  \underbrace{\mathbf{g}(\mathbf{x}(x_1), \mathbf{x}(x_2), \ldots, \mathbf {x}(x_m))}_{\text{branch}}, \underbrace{\mathbf{f}(\zeta)}_{\text{trunk}}  \rangle \Bigg\rvert < \epsilon \]
holds for all $\mathbf{x} \in V$ and $\zeta \in K_2$, where $\langle \cdot, \cdot \rangle$ denotes the dot product in $\mathbb{R}^p$. For the two functions $\mathbf{g}, \mathbf{f}$ classical deep neural network models and architectures can be chosen that satisfy the universal approximation theorem of functions, such as fully connected networks or convolutional neural networks.
\end{theorem}
\noindent The interested reader can find more information and details regarding the proof of the theorem in \cite{lu2021learning}.

\section{Batching details of separable physics-informed DeepONet}
\label{sec:architecture_sep_PI_DeepONET} 

\begin{figure}[t!]
    \includegraphics[trim= 120 430 120 80, clip, width=\textwidth]{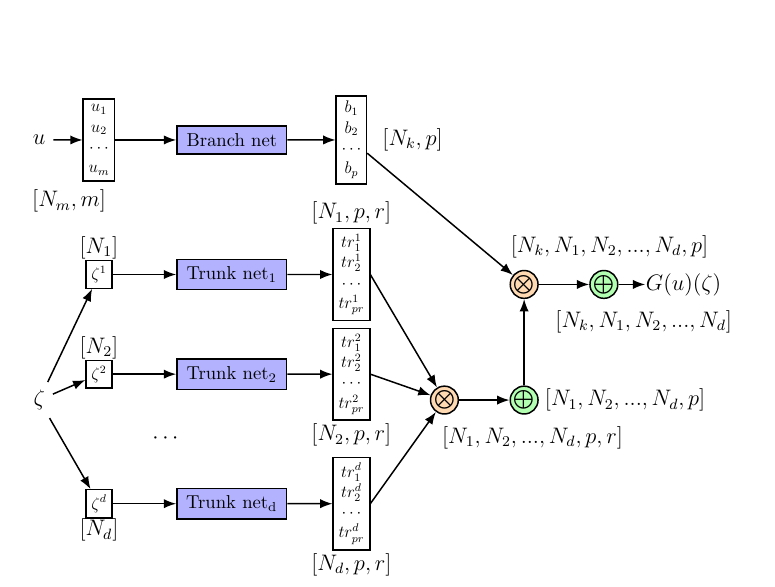}
    \caption{Separable PI-DeepONet Architecture: This figure illustrates the Separable PI-DeepONet architecture, which employs batch-based forward passes with $N_b$ input functions sampled at $m$ sensors. The $d$ input coordinates for $y$ are processed as factorized pairs, with a unique network per trunk input dimension. The branch network, consistent with classical DeepONet, outputs a hidden dimension $p$. Each of the $d$ trunk networks produces a tensor of length $p \cdot r$. The architecture's key operations include: Initial outer product $\bigotimes$ in the trunk over batches in each network; Summation $\bigoplus$ over tensor rank $r$; Second outer product combining branch and trunk batches; Final summation over hidden dimension $p$.}
    \label{fig:sep_framework}
\end{figure}

In algorithm \ref{alg:schematic}, we have presented a flow of the operations for the Sep-DeepONet framework. The algorithm indicates the evolution of the matrix sizes during each step. The same problem is visualized in figure \ref{fig:sep_framework}.

\begin{algorithm}[H]
\SetAlgoLined
\caption{Separable DeepONet for a general problem with $d$ trunk networks.}
\label{alg:schematic}
 \KwIn{
   Coordinates:    $(\zeta^{\gamma}_{\delta})_{\delta=1}^{N_\gamma}, \quad\gamma \in \{i,j,\dots d\}$,\\
    Input Functions: $(u^{k})_{k=1}^{N_k}$ evaluated at $m$ sensor locations $(x_n)_{n=1}^m$,\\
    Parameters: $p$ (hidden dim), $r$ (tensor rank)}
\KwOut{$\mathcal{G}(u)_{mij\dots d}$}

\textbf{Step 1:} Process Trunk Networks\;
\For{$\gamma \leftarrow 1$ \KwTo $d$}{
\For{$\delta \leftarrow 1$ \KwTo $N_{\gamma}$}{
    Compute $tr^{\gamma}_{\delta\alpha\beta} = \text{TrunkNet}_\gamma(\zeta^{\gamma}_{\delta})$\;}}
\textbf{Step 2:} Process Branch Network\;
\For{$k \leftarrow 1$ \KwTo $N_k$}{
    Compute $br_{k\alpha} = \text{BranchNet}(u^{k})$\;}
\textbf{Step 3:} Combine Trunk Outputs\;
$tr_{ij\dots d\alpha}=\sum_{\beta=1}^{r}tr^{1}_{i\alpha\beta}\cdot tr^{2}_{j\alpha\beta}\dots tr^{d}_{d\alpha\beta}$ = einsum(`i$\alpha\beta$,j$\alpha\beta$,$\dots$,d$\alpha\beta\rightarrow$ij\dots d$\alpha$', $tr^1$, $tr^2$, \dots, $tr^d$)\;
\textbf{Step 4:} Compute Final Output\;
$\mathcal{G}(u)_{kij\dots d} = \sum_{\alpha=1}^{p} br_{k\alpha} \cdot tr_{ij\dots d\alpha}$ = einsum(`k$\alpha$, ij\dots d$\alpha\rightarrow$ kij\dots d', $br$, $tr$)\;
\end{algorithm}

\end{document}